%% file: main.tex
\documentclass[sigconf]{acmart}

\AtBeginDocument{%
  \providecommand\BibTeX{{%
    \normalfont B\kern-0.5em{\scshape i\kern-0.25em b}\kern-0.8em\TeX}}}

\setcopyright{acmcopyright}
\copyrightyear{2023}
\acmYear{2023}
\acmDOI{10.1145/3637528.3671815}

\copyrightyear{2024}
\acmYear{2024}
\setcopyright{acmlicensed}\acmConference[KDD '24]{30th ACM SIGKDD Conference on Knowledge Discovery and Data Mining }{August 25--29, 2024}{Barcelona, Spain}
\acmBooktitle{30th ACM SIGKDD Conference on Knowledge Discovery and Data Mining (KDD '24), August 25--29, 2024, 2024, Barcelona, Spain}
\acmISBN{979-8-4007-0490-1/24/08}
\acmDOI{10.1145/3637528.3671815}
\usepackage{tikz}
\usetikzlibrary{arrows.meta}

\usepackage{mathtools}

\usepackage[normalem]{ulem}
\usepackage{multirow}
\usepackage{float}
\usepackage{enumitem}
\usepackage{multicol}
\usepackage{chemfig}
\usepackage{caption}
\usepackage[font=small]{subfig}
\usepackage{xspace}

\usepackage{algorithm}
\usepackage[noend]{algorithmic}

\setlist{nolistsep,leftmargin=*}

\usepackage{booktabs}
\usepackage{multirow,comment}

\input{preamble}

\begin{document}

\title{\namemodel: A Neural Approach for Robust Graph Partitioning}

\author{Rishi Shah}
\authornote{These authors contributed equally to this research.}

\affiliation{%
  \institution{Department of Computer Science and Engineering, IIT Delhi, India}
  \country{}
}\email{rishi.shah10122001@gmail.com}

\author{Krishnanshu Jain}\authornotemark[1]
\affiliation{%
  \institution{Department of Computer Science and Engineering, IIT Delhi, India}
  \country{}
}\email{krishnanshu1907@gmail.com}

\author{Sahil Manchanda}\authornotemark[1]
\affiliation{%
  \institution{Department of Computer Science and Engineering, IIT Delhi, India}
\country{}
}\email{sahilm1992@gmail.com}

\author{Sourav Medya}
\affiliation{%
 \institution{University of Illinois, Chicago, USA}
\country{}
}\email{medya@uic.edu}

\author{Sayan Ranu}
\affiliation{%
  \institution{Department of Computer Science and Engineering, IIT Delhi, India}
\country{}
  }\email{sayanranu@iitd.ac.in}

\renewcommand{\shortauthors}{Rishi Shah, Krishnanshu Jain, Sahil Manchanda, Sourav Medya, \& Sayan Ranu}


\input{01_abstract}

\ccsdesc[500]{Computing methodologies~Machine learning}
\ccsdesc[500]{Computing methodologies~Artificial intelligence}
\ccsdesc[500]{Mathematics of computing~Graph algorithms}
\ccsdesc[500]{Mathematics of computing~Graph algorithms}

\keywords{ Graph Partitioning, Min Cut, Robustness, Versatile Partitioning Objectives, Inductive learning, GNN, Reinforcement learning. }

\maketitle

\input{02_Introduction}

\input{03_problem}

\input{04_method}

\input{05_expts}

\input{06_Conclusion}

\input{08_Ack}
\clearpage
\bibliographystyle{ACM-Reference-Format}
\bibliography{www24}

\appendix
\input{07_appendix}

\end{document}

%% file: preamble.tex
\newcommand{\gnn}{\textsc{Gnn}{}}

\newcommand{\namemodel}{\textsc{NeuroCUT}{}}
\newcommand{\dmon}{\textsc{DMon}{}}
\newcommand{\gap}{\textsc{GAP}{}}
\newcommand{\kmeans}{\textsc{K-means}{}}
\newcommand{\ortho}{\textsc{Ortho}{}}
\newcommand{\mcpool}{\textsc{MinCutPool}{}}

\newcommand{\CG}{\mathcal{G}\xspace}

\newcommand{\CT}{\mathcal{T}\xspace}

\newcommand{\CP}{\mathcal{P}\xspace}
\newcommand{\CV}{\mathcal{V}\xspace}
\newcommand{\CE}{\mathcal{E}\xspace}
\newcommand{\CC}{\mathcal{C}\xspace}

\newcommand{\CS}{\mathcal{S}\xspace}

\newcommand{\Obj}{{Obj}}

\newtheorem{defn}{\textbf{Definition}}

\newtheorem{prob}{\textbf{Problem}}

\newcommand{\rev}[1]{\textcolor{black}{#1}}

\usepackage{colortbl}

\usepackage{amsmath,scalerel}
\usepackage{xcolor}
\usepackage{graphicx}

\usepackage{caption}



%% file: 01_abstract.tex
\begin{abstract}

Graph partitioning aims to divide a graph into $k$ disjoint subsets while optimizing a specific partitioning objective. The majority of formulations related to graph partitioning exhibit NP-hardness due to their combinatorial nature. Conventional methods, like approximation algorithms or heuristics, are designed for distinct partitioning objectives and fail to achieve generalization across other important partitioning objectives. Recently machine learning-based methods have been developed that learn directly from data. Further, these methods have a distinct advantage of utilizing node features that carry additional information. However, these methods assume differentiability of target partitioning objective functions and cannot generalize for an unknown number of partitions, i.e., they assume the number of partitions is provided in advance. In this study, we develop {\namemodel} with two key innovations over previous methodologies. First, by leveraging a reinforcement learning-based framework over node representations derived from a graph neural network and positional features, {\namemodel} can accommodate \textit{any} optimization objective, even those with non-differentiable functions. Second, we decouple the parameter space and the partition count making {\namemodel}  \textit{inductive} to any unseen number of partition, which is provided at query time. Through empirical evaluation, we demonstrate that {\namemodel} excels in identifying high-quality partitions, showcases strong generalization across a wide spectrum of partitioning objectives, and exhibits strong generalization to unseen partition count.

\end{abstract}

%% file: 02_Introduction.tex
\section{Introduction}

Graph partitioning is a fundamental problem in network analysis with numerous real-world applications in various domains such as system design in online social networks \cite{pujol2009divide}, dynamic ride-sharing in transportation systems  \cite{tafreshian2020trip}, VLSI design~\cite{kahng2011vlsi}, and preventing cascading failure in power grids \cite{li2005strategic}. The goal of graph partitioning is to divide a given graph into disjoint subsets where nodes within each subset exhibit strong internal connections while having limited connections with nodes in other subsets. Generally speaking, the aim is to find somewhat balanced partitions while minimizing the number of
edges across partitions.

\subsection{Related Work}

Several graph partitioning formulations have been studied in the literature, mostly in the form of discrete  optimization~\cite{karypis1997multilevel,karypis1998multilevelk,karypis1999multilevel,andersen2006local,chung2007four}. The majority of the formulations are NP-hard and thus the proposed solutions are either heuristics or algorithms with approximate solutions \cite{karypis1997multilevel}. \rev{Among these, two widely used methods are Spectral Clustering~\cite{ng2001spectral} and hMETIS~\cite{karypis1997multilevel}. Spectral Clustering partitions a graph into clusters based on the eigenvectors of a similarity matrix derived from the graph. hMETIS is a hypergraph partitioning method that divides a graph into clusters by maximizing intra-cluster similarity while minimizing inter-cluster similarity. However, such techniques are confined to specific objective functions and are unable to leverage the available node features in the graph.} 
Recently, there have been attempts to solve graph partitioning problem via neural approaches. The neural approaches have a distinct advantage that they can utilize node features. Node features supply additional information and provide contextual insights that may improve the accuracy of graph partitioning. For instance, \cite{wang2017mgae} has recognized the importance of incorporating node attributes in graph clustering where nodes are partitioned into disjoint groups. Note that there are existing neural approaches~\cite{khalil2017learning, gcomb, kool2018attention, manchanda2022generalization,manchanda2024generative, joshi2019efficient,tian2024combhelper, gasse2019exact,ranjan2022greed, manchanda2023limip, graphgen} to solve other NP-hard graph combinatorial problems (e.g., minimum vertex cover). However these methods are not generic enough to solve graph partitioning. \rev{For example, although S2VDQN~\cite{khalil2017learning} learns to solve the Maxcut problem, it is designed for the specific case of bi-partitioning. This hinders its applicability to the target problem setup of $k$-way partitioning\footnote{Additional related work is present in  Appendix \ref{app:additional_rel_work}}.}  

\looseness=-1 

In this paper, we build a single framework to solve several graph partitioning problems. One of the most relevant to our work is the method DMoN by \citet{muller2023graph}. This method designs a neural architecture for cluster assignments and use a modularity-based objective function for optimizing these assignments. Another method, that is relevant to our work is GAP, which is an unsupervised learning method to solve the balanced graph partitioning problem \cite{nazi2019gap}. It proposes a differentiable loss function for partitioning based on a continuous relaxation of the normalized cut formulation.  Deep-MinCut being an unsupervised approach learns both node embeddings and the community structures simultaneously where the objective is to minimize the mincut loss \cite{duong2023deep}. Another method solves the multicut problem where the number of partitions is \textit{not} an input to the problem \cite{jung2022learning}. The idea is to construct a reformulation of the multicut ILP constraints to a polynomial program as a loss function. However, the problem formulation is different than the generic normalized cut or mincut problem.  Another related work is DGCluster~\cite{bhowmick2024dgcluster}, which  proposes to solve the attributed graph clustering problem while maximizing modularity when $k$ is not known beforehand.  Finally, ~\cite{gatti2022graph} solves the normalized cut problem only for the case where the number of partitions is exactly two. Nevertheless, these neural approaches for the graph partitioning problem often do not use node features and only limited to a distinct partitioning objective. Here, we point out notable drawbacks that we address in our framework.

\begin{itemize}
    \item \textbf{Non-inductivity to partition count:} The number of partitions required to segment a graph is an input parameter. Hence, it is important for a learned model to generalize to any partition count without retraining. \rev{In $k$-way graph partitioning, a model demonstrates inductivity to the number of partitions when it can infer on varying partition numbers without specific training for each.} Existing neural approaches are non-inductive to the number of partitions, i.e they can only infer on number of partitions on which they are trained. Additionally, it is worth noting that the optimal number of partitions is often unknown beforehand. \rev{This capability is crucial in practical applications like chip design, where graph partitioning optimizes logic cell placement by dividing netlists (circuits) into smaller subgraphs, aiding independent placement. As the optimal partition count is frequently unknown in advance, experimenting with different partition numbers is a common practice.} Hence, it is a common practice to experiment with different partition counts and evaluate their impact on the partitioning objective. While the existing methods \cite{muller2023graph,nazi2019gap,gatti2022graph} assume that the number of partitions ($k$) is known beforehand, our proposed method can generalize to any $k$. 
    
    \item \textbf{Non-generalizability to different cut functions:} Multiple objective functions for graph cut have been studied in the partitioning literature. The optimal objective function hinges upon the subsequent application in question. For instance, the two most relevant studies, DMoN \cite{muller2023graph} and GAP \cite{nazi2019gap} focus on maximizing modularity and minimizing normalized cut respectively. Our framework is generic and can solve different partitioning objectives. 
    
  \item {\bf Assumption of differential objective function: } Existing neural approaches assume the objective function to be differentiable. As we illustrate in \S~\ref{sec:formulation}, the assumption does not always hold in the real-world. For instance, the sparsest cut \cite{chawla2006hardness} and balanced cut \cite{nazi2019gap} objectives are not differentiable.

\end{itemize}

\subsection{Our Contributions}
In this paper, we circumvent the above-mentioned limitations through the following key contributions.

\begin{itemize}
    \item \textbf{Versatile objectives:}  We develop {\namemodel}; an auto-regressive, graph reinforcement learning framework that integrates positional information, to solve the graph partitioning problem for attributed graphs. Diverging from conventional algorithms, {\namemodel} can solve multiple partitioning objectives. Moreover, unlike other neural methods, {\namemodel} can accommodate diverse partitioning objectives, without the necessity for differentiability.
    \looseness=-1

\item \textbf{Inductivity to  number of partitions:} The parameter space of {\namemodel} is independent of the partition count. This innovative decoupled architecture endows {\namemodel} with the ability to generalize effectively to arbitrary partition count specified during inference.
    \looseness=-1

 \item \textbf{Empirical Assessment:} We perform comprehensive experiments employing real-world datasets, evaluating {\namemodel} across four distinct graph partitioning objectives. Our empirical investigation substantiates the efficacy of {\namemodel} in partitioning tasks, showcasing its robustness across a spectrum of objective functions. We also demonstrate the capability of {\namemodel} to generalize effectively to partition sizes that have not been seen during training.

\end{itemize}


%% file: 03_problem.tex
\section{Problem Formulation}
\label{sec:formulation}

In this section, we introduce the concepts central to our work and formulate the problem. All the notations used in this work are outlined in Table~\ref{tab:notation}.\vspace{-0.03in}

\begin{defn}[Graph]    

We denote a graph as $\mathcal{G}=(\mathcal{V}, \mathcal{E}, \textbf{{X}})$ where  $\CV $ is the set of nodes, $\CE \subseteq \CV \times \CV $ is the set of edges and $\textbf{{X}} \in \mathbb{R}^{|\mathcal{V}|\times |F|}$ refers to  node feature matrix where $F$ is the set of all features in graph $\mathcal{G}$.



\end{defn}

\begin{defn}[Cut]
A cut $\CC=(\CS, \CT)$ is a partition of $\CV$ into two subsets $\CS$ and $\CT$. 
The cut-set of $\CC = (\CS,\CT)$ is the set $\rev{\{}(u,v) \in \CE | u \in \CS, v \in \CT\rev{\}}$ of edges that have one endpoint in $\CS$ and the other endpoint in $\CT$.

\end{defn}

\begin{defn}[Graph Partitioning]
Given graph $\CG$, we aim to partition $\CG$ into $k$ disjoint sets $\CP = \{P_1, P_2,\cdots, P_k \}$ such that the union of the nodes in those sets is equal to $\CV$
 i.e $\cup_{i=1}^{k}P_{i}=\CV $ and each node belongs to exactly one partition.
 \label{def:graph_part}
\end{defn}

\noindent \label{sec:partitioning_obj}
\textbf{ Partitioning Objective:} We aim to minimize/maximize a partitioning objective of the form $\Obj(\CG, \CP)$. A wide variety of objectives for graph partitioning have been proposed in the literature. Without loss of generality, we consider the following four objectives. These objectives are chosen due to being well studied in the literature, while also being diverse from each other. \footnote{Our framework is not restricted to these objectives.}
\begin{enumerate}
    \item \textit{$k$-MinCut \cite{saran1995finding}:} Partition a graph into $k$ partitions such that the total number of edges across partitions is minimized.\vspace{-0.07in} \begin{equation}
    \text{$k$-mincut}(\mathcal{P})=\sum_{l=1}^{|\mathcal{P}|} \frac{|\operatorname{cut}\left(P_l , \overline{P_l} \right)|}{\sum_{e \in \CE} |e|}
    \end{equation}    
    Here $P_l$ refers to the set of  elements in $l^{th}$ partition of $\CP$ as described in Def. ~\ref{def:graph_part} and $\overline{P_l}$ refers to set of elements not in $P_l$. 
    \item \textit{Normalized Cut \cite{shi2000normalized}:} The $k$-MinCut criteria favors cutting small sets of isolated nodes in the graph. To avoid this unnatural bias for partitioning out small sets of points, normalised cut computes the cut cost as a fraction of the total edge connections to all the nodes in the graph.\vspace{-0.05in} \begin{equation}
    \operatorname{Ncut}(\mathcal{\CP}):=\sum_{l=1}^{|\CP|} \frac{|\operatorname{cut}\left(P_l, \overline{P_l} \right)|}{\operatorname{vol}\left(P_l, \CV \right)}  \end{equation}
    Here, $\operatorname{vol}\left(P_l, \CV \right) := \sum_{v_i \in P_l, v_j \in \CV} e(v_i, v_j)  $. 
    \item \textit{Balanced Cut \cite{nazi2019gap}:} Balanced cut favours partitions of equal sizes so an extra term that indicates the squared distance from equal sized partition is added to normalized cuts.\begin{equation}
    \label{eq:balanced_cuts}
        \operatorname{Balanced-Cuts}(\mathcal{\CP}):=\sum_{l=1}^{|\CP|} \frac{|\operatorname{cut}\left(P_l, \overline{P_l} \right)|}{\operatorname{vol}\left(P_l, \CV \right)} + \frac{(|P_l| - |\CV|/k)^2}{|\CV|^2}
    \end{equation}
    Here, $\operatorname{vol}\left(P_l, \CV \right) := \sum_{v_i \in P_l, v_j \in \CV} e(v_i, v_j)  $
    \item \looseness=-1  Sparsest Cut \cite{chawla2006hardness}: Two-way sparsest cuts minimize the cut edges relative to the number of nodes in the smaller partition. We generalize it to $k$-way sparsest cuts by summing up the value for all the partitions. The intuition behind sparsest cuts is that any partition should neither be very large nor very small.
    \begin{alignat}{2}
     \hspace*{-0.1in}\operatorname{Sparsest-Cuts}(\mathcal{\CP})&=\sum_{l=1}^{|\CP|}\phi(P_l,\overline{P_l}) \\
     where\;\; \phi(S, \bar{S})&=\frac{\operatorname{cut}(S, \bar{S})}{\min (|S|,|\bar{S}|)}
    \end{alignat}
\end{enumerate}
\vspace{-0.05in}

\begin{prob}[Learning to Partition Graph ]
 Given a graph $\CG$ and the number of partitions $k$, the goal is to find a partitioning $\CP$ of the graph $\CG$ that optimizes a target objective function $\Obj(\CG,\CP)$. Towards this end, we aim to learn a policy $\pi$ that assigns each node $v \in \CV$ to a partition in $\CP$.
 \end{prob}

In addition to our primary goal of finding a partitioning that optimizes a certain objective function, we also desire policy $\pi$ to have the following properties:

\begin{enumerate}
    \item \textbf{Inductive:} Policy $\pi$ is inductive if the parameters of the policy are independent of both the size of the graph and the number of partitions $k$.
    If the policy is not inductive then it will be unable to infer on unseen size graphs/number of partitions.

    \item \textbf{Learning Versatile Objectives:} 
    To optimize the parameters of the policy,   a target objective function is required. The optimization objective may not be differentiable and it might not be always possible to obtain a differentiable formulation. Hence, the policy $\pi$ should be capable of learning to optimize for a  target objective that may or may not be differentiable. 
\end{enumerate}

 \begin{figure*}[h!]
    \centering
    \includegraphics[scale=0.35]{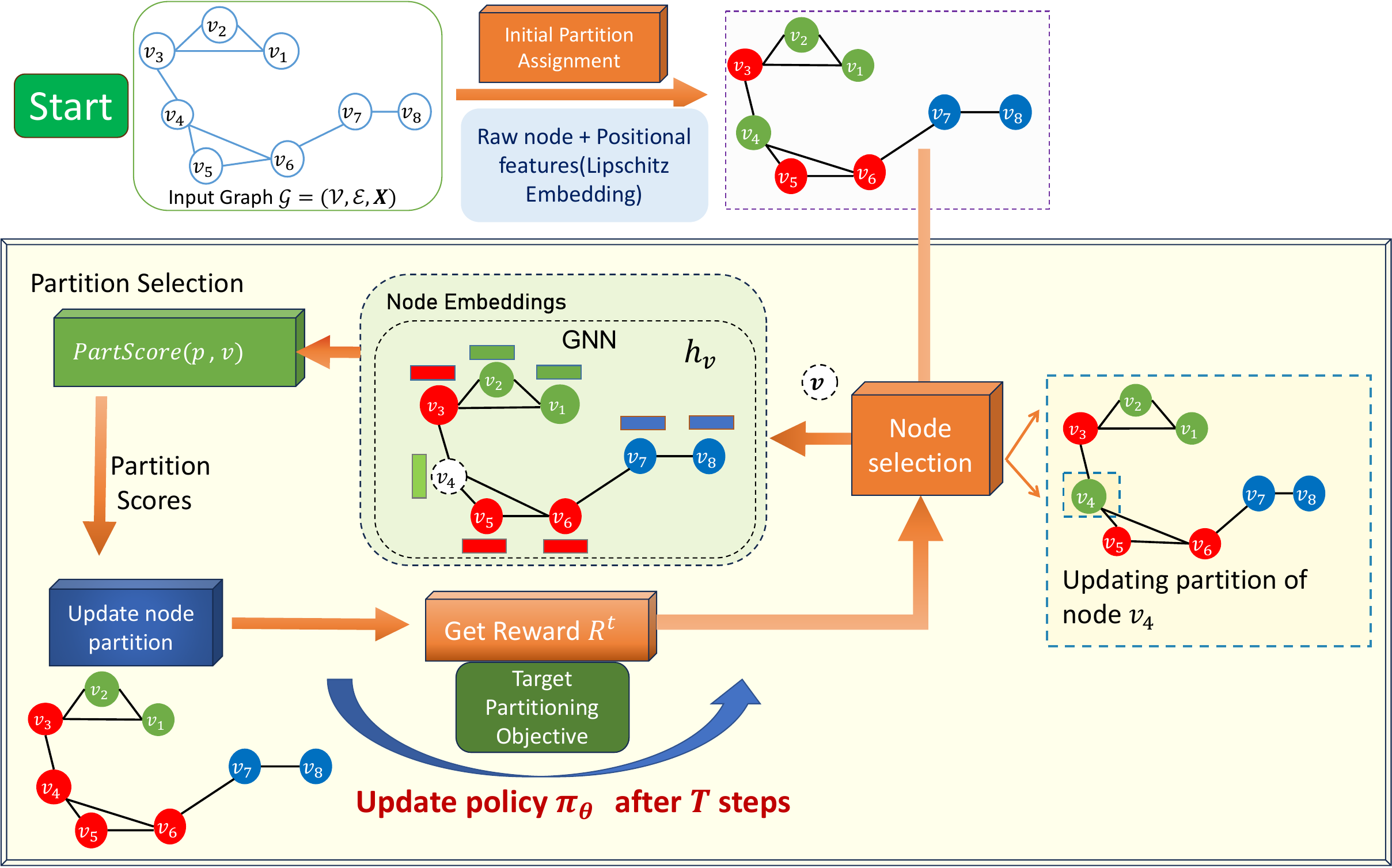}
    \vspace{-0.13in}
    \caption{Architecture of {\namemodel}. First, the initial partitioning of the graph is performed based on node features and positional embeddings. These embeddings are refined using GNN to infuse toplogical information from neighborhood. At each step a node is selected and its partitioned is updated. During training the GNN parameters are updated and hence embeddings are re-computed. During inference, the GNN is called only once to compute the embeddings of the nodes of the graph. }
    \label{fig:arch}
    \vspace{-0.20in}
\end{figure*}

%% file: 04_method.tex
\begin{table}[t!]
    \centering
    \caption{Notations used in the paper}
    \vspace{-0.1in}
    \scalebox{0.95}{
        \begin{tabular}{m{0.2 \linewidth}  m{0.75 \linewidth}}
         \toprule
         \textbf{Symbol} & {\textbf{Meaning}} \\ 
         \midrule
         $\mathcal{G}$ & Graph \\ \cmidrule(lr){1-2}
         $\mathcal{V}$& Node set \\ \cmidrule(lr){1-2}
          $e$ & Edge $e \in \mathcal{E}$ \\ \cmidrule(lr){1-2}
         $\mathcal{E}$& Edge set \\ \cmidrule(lr){1-2}
    $\textbf{\textit{X}}$ & Feature matrix containing raw node features \\ \cmidrule(lr){1-2}
$\mathcal{N}_v$&Neighboring nodes of node $v$ \\ \cmidrule(lr){1-2}

     $Obj(\CG, \CP) $ &  Objective function based upon graph $\CG$ and its partitioning $\CP$   \\ \cmidrule(lr){1-2}

      $k$ & Number of partitions   \\ \cmidrule(lr){1-2}
      
           $\mathcal{P}^t$& Partitioning at time $t$\\ \cmidrule(lr){1-2}
           
           ${P}^t_i$& $i^{th}$ partition  at time $t$\\ \cmidrule(lr){1-2}

           $\overline{{P}_i}$& Set of nodes that are not in the $i^{th}$ partition \\ \cmidrule(lr){1-2}

         $\mathcal{S}^t$& State of system at step $t$\\ \cmidrule(lr){1-2}

          $\operatorname{PART}(\CP^t,v)$& Partition of node $v$ at time $t$\\ \cmidrule(lr){1-2}
         
      $\mathcal{S}^t$& State representation of Partitions and Graph at step $t$\\ \cmidrule(lr){1-2}

         $\mathbf{pos}(v)$ & Positional embedding of node $v$   \\ \cmidrule(lr){1-2}

                  $\mathbf{emb_{init}}(v)$ & Initial  embedding of node $v$   \\ \cmidrule(lr){1-2}

 $\alpha$ & Number of anchor nodes for lipschitz embedding   \\ \cmidrule(lr){1-2}
      
         $T$ & Length of trajectory \\ \cmidrule(lr){1-2}    
         
         $\pi$ & Policy function\\ \cmidrule(lr){1-2}
        \bottomrule
        \end{tabular}
        }
            
    \label{tab:notation}
\end{table}

\section{\namemodel: Proposed Methodology}
\label{sec:neurocut}
Fig.~\ref{fig:arch} describes the framework of \namemodel. For a given input graph $\CG$, we first construct the initial partitions of nodes using a clustering based approach. Subsequently, a message-passing {\gnn} embeds the nodes of the graph ensuring inductivity to different graph sizes. Next, the assignment of nodes to partitions proceeds in a two-phased strategy. We first select a node to change its partition and then we choose a suitable partition for the selected node. Further, to ensure inductivity on the number of partitions, we \textit{decouple} the number of partitions from the direct output representations of the model. This decoupling allows us to query the model to an unseen number of partitions. After a node's partition is updated, the \textit{reward} with respect to change in partitioning objective value is computed and the parameters of the policy are optimized. The reward is learned through \textit{reinforcement learning (RL)}~\cite{sutton2018reinforcement}. 

\looseness=-1 The choice of using RL is motivated through two observations. Firstly, cut problems on graphs are generally recognized as NP-hard, making it impractical to rely on ground-truth data, which would be computationally infeasible to obtain. Secondly, the cut objective may lack differentiability. Therefore, it becomes essential to adopt a learning paradigm that can be trained even under these non-differentiable constraints. In this context, RL effectively addresses both of these critical requirements. Additionally, in the process of sequentially constructing a solution, RL allows us to model the gain obtained by perturbing the partition of a node.

\looseness=-1

\noindent
\textbf{Markov Decision Process. }Given a graph $\CG$, our objective is to find the partitioning $\CP$ that maximizes/minimizes the target objective function $Obj(\CG, \CP)$.  We model the task of iteratively updating the partition for a node as a \textit{Markov Decision Process (MDP)} defined by the tuple $(S,\mathcal{A},\rho,R,\gamma)$. Here, $S$ is the \textit{state space}, $\mathcal{A}$ is the set of all possible \textit{actions}, $\rho:S\times S \times \mathcal{A} \rightarrow [0,1]$  denotes the \textit{state transition probability function},  $R:S\times \mathcal{A} \rightarrow \mathbb{R}$ denotes the \textit{reward function} and $\gamma \in (0,1)$ the \textit{discounting factor.}  We next formalize each of these notions in our MDP formulation.

\subsection{State: Initialization \& Positional Encoding}
\label{sec:initialization}
\noindent \textbf{Initialization.} Instead of directly starting from empty partitions, we perform a warm start operation that clusters the nodes of the graph to obtain the initial graph partitions. The graph is clustered into $k$ clusters based on their raw features and positional embeddings (discussed below). We use \textit{K-means}~\cite{macqueen1967some} algorithm for this task. The details of the clustering are present in Appendix~\ref{app:clustering}. 

\noindent \\ 
\textbf{Positional Encoding (Embeddings).} 
    Given that the partitioning objectives are NP-hard mainly due to the combinatorial nature of the graph structure, we look for representations that capture the location of a node in the graph. Positional encodings provide an idea of the position in space of a given node within the graph. Two nodes that are closer in the graph, should be closer in the embedding space. Towards this, we use \textit{Lipschitz Embedding}~\cite{bourgain1985lipschitz}. Let $\mathcal{A}=\{a_1,\cdots,a_\alpha\}\subseteq\CV$ be a randomly selected subset of $\alpha$ nodes. We call them anchor nodes. From each anchor node $i$, the walker starts a random walk \cite{bianchini2005inside} and jumps to a neighboring node $j$ with a transition probability ($\mathbf{W}_{ij}$) governed  by the transition probability matrix $\mathbf{W} \in \mathbb{R}^{|V| \times |V|}$. Furthermore, at each step, with probability $c$ the walker jumps to a neighboring node $j$ and returns to the node $i$ with $1-c$. Let $\vec{r}_{ij}$ corresponds to the probability of the random walker starting from node $i$ and reaching node $j$. \vspace{-0.05in}
\begin{equation}
\label{eq:RWR_lipschitz}
\vec{r}_i=c \tilde{\mathbf{W}} \vec{r}_i+(1-c) \vec{e}_i
\end{equation} 
Eq.~\ref{eq:RWR_lipschitz} describes the random walk starting at node $i$. In vector $\vec{e}_i \in \mathbb{R}^{|V| \times 1}$, only the $i^{th}$ element (the initial anchor node) is $1$, and the rest are set to zero. We set $\mathbf{W}_{ij}=\frac{1}{degree(j)}$ if edge $e_{ij} \in \CE$, $0$ otherwise. 
The random walk with restart process is repeated for $\beta$ iterations, where $\beta$ is a hyper-parameter.  Here $\vec{e}_i  \in \mathbb{R}^{|V| \times 1 } $ and $c$ is a scalar. 

Based upon the obtained random walk vectors for the set of anchor nodes $\mathcal{A}$, we embed all nodes  $u \in \CV$ in a $\alpha$-dimensional feature space: \vspace{-0.05in}
\begin{equation}
 \mathbf{pos}(u)=[r_{1u}, r_{2u}, \cdots, r_{\alpha u}]   
\end{equation}

To accommodate both raw feature and positional information we concatenate the raw node features i.e $\textbf{X}[u]$ with the positional embedding $\mathbf{pos}(u)$ for each node $u$ to obtain the \textit{initial embedding} which will act as input to our neural model. Specifically,
\vspace{-.04in}
\begin{equation}
\label{eq:emb_init}
 \mathbf{emb_{init}}(u)=  \textbf{X}[u]   \;\; \mathbin{\|} \;\; \mathbf{pos}(u) 
\end{equation}

In the above equation, $\mathbin{\|}$ represents the concatenation operator.

\noindent \\
\textbf{State.}
The state space characterizes the state of the system at time $t$ in terms of the current set of partitions $\CP^t$. Intuitively the state should contain information to help our model make a decision to select the next node and the partition for the node to be assigned. 
 Let $\CP^t$ denote the status of partitions at time $t$ wherein a partition  $P^t_i$ is represented by all nodes belonging to $i^{th}$ partition.  The state of the system at step $t$  is defined as 
 \begin{equation}
 S^t = \{  S_1^t, S_2^t, \cdots S_k^t : \; S_i^t{=}\{\mathbf{emb_{init}}(v) \; \;  \forall v \in P_i^t \}  \}
 \label{eq:state}
 \end{equation}
   Here state of each partition is represented by the collection of  \textit{initial embedding} of nodes in that partition.
\subsection{Action: \textit{Selection of Nodes} and \textit{Partitions} }
\noindent
 Towards finding the best partitioning scheme for the target partition objective we propose a \textbf{\textit{2-step action}} strategy to update the partitions of nodes. The first phase consists of identifying a node to update its partition. Instead of arbitrarily picking a node, we propose to prioritize selecting nodes for which the new assignment is more likely to improve the overall partitioning objective.  In comparison to a strategy that arbitrarily selects nodes, the above mechanism promises greater improvement in the objective with less number of iterations. In the second phase, we calculate the score of each partition $\CP$ with respect to the selected node from the first phase and then assign it to one of the partitions based upon the partitioning scores. We discuss both these phases in details below.

\subsubsection{\textbf{Phase 1: Node Selection to Identify Node to Perturb}}
\label{sec:phase1_node_selection}
\noindent \\
Let $\operatorname{PART}(\CP^t,v)$ denote the partition of the node $v$ at step $t$. Our proposed formulation involves selecting a node $v$ at step $t$ belonging to partition $\operatorname{PART}(\CP^t,v)$ and then assigning it to a new partition. The newly assigned partition and the current partition of the node could also be same.

Towards this, we design a heuristic to prioritize selecting nodes which when placed in a new partition are more likely to improve the overall objective value. A node $v$ is highly likely to be moved from its current partition if most of its neighbours are in a different partition than that of node $v$. Towards this, we calculate the score of nodes $v \in \CV$ in the graph as the ratio between the maximum number of  neighbors in another partition and the number of neighbors in the same partition as $v$. Intuitively, if a partition exists in which the majority of neighboring nodes of a given node $v$ belong, and yet node $v$ is not included in that partition, then there is a high probability that node $v$ should be subjected to perturbation. Specifically the score of node $v$ at step $t$ is defined as: \vspace{-0.13in}

\begin{multline}\label{eqn:node_selection}
\operatorname{NodeScore}^t[v] = \\
  \frac{\max_{p \in {P}^t \setminus \operatorname{PART}({P}^t,v)} 
  { \vert u| u \in \mathcal{N}(v) \ni \operatorname{PART}({P}^t,u) = p \vert}}
  { \vert u|u \in \mathcal{N}(v) \ni \operatorname{PART}({P}^t,u) = \operatorname{PART}({P}^t,v) \vert}
  \times \frac{1}{degree(v)}
\end{multline}

\rev{For a node of interest $v$, the numerator computes the maximum number of neighbors in a different partition than that of $v$. As described in Table ~\ref{tab:notation}, $\operatorname{PART}(P^t,v)$ refers to the partition of $v$ at step $t$. The expression $P^t \setminus \operatorname{PART}(P^t,v)$ computes all other partitions except the partition of $v$. The term $\vert u| u \in \mathcal{N}(v) \ni \operatorname{PART}(P^t,u) = p \vert$ computes the number of neighbors of $v$ in the partition $p$. The denominator computes the number of neighbors of node $v$ in the same partition as $v$, referred to as $\operatorname{PART}(P^t,v)$. }
         Further, a node having a higher degree implies that it has several edges associated to it. Hence, an incorrect placement of it could contribute to a higher partitioning value. Therefore, we normalize the scores by the $degree$ of the node. 
\subsubsection{\textbf{Phase 2: Inductive Method for Partition Selection}}
\label{sec:phase2_node_selection}
\noindent \\
Once a node is selected, the next phase involves choosing the new partition for the node. Towards this, we design an approach empowered by Graph Neural Networks (GNNs)~\cite{hamilton2017inductive} which enables the model to be inductive with respect to size of graph. Further, instead of predicting a fixed-size score vector~\cite{nazi2019gap, tsitsulin2023graph} for the number of partitions, our proposed method of computing partition scores allows the model to be inductive to the number of partitions too. We discuss both above points in section below. 

\noindent \\
\textbf{\textit{Message Passing through Graph Neural Network}}

\noindent
To capture the interaction between different nodes and their features along with the graph topology, we parameterize our policy by a Graph Neural Network (GNN)~\cite{hamilton2017inductive}. GNNs combine node feature information and the graph structure to learn better representations via feature propagation and aggregation. 

We first initialize the input layer of each node $u \in \CV$ in graph as $\mathbf{h}^0_u= \mathbf{emb_{init}}(u)$ using eq.~\ref{eq:emb_init}. We perform $L$ layers of message passing to compute representations of nodes.  To generate the embedding for  node $u$ at layer $l+1$ we perform the following transformation\cite{hamilton2017inductive}: \vspace{-0.05in}
\begin{equation}
\hspace{0.55in}
  \label{eq:gnn}
  \begin{aligned}
  \mathbf{h}_u^{l+1} &= \mathbf{W}^l_1 \mathbf{h}_u^{l} + \mathbf{W}^l_2 \cdot \frac{1}{\vert \mathcal{N}_u \vert} \sum_{u' \in  \mathcal{N}_{u} } \mathbf{h}_{u'}^l
  \end{aligned}\vspace{-0.05in}
\end{equation}

where $\mathbf{h}_u^{(l)}$ is the node embedding in layer $l$. $\mathbf{W}^l_1$  and $\mathbf{W}^l_2$  are trainable weight matrices at layer $l$.

Following $L$ layers of message passing, the final node representation of node $u$ in the $L^{th}$ layer is denoted by $\mathbf{h}_u^L \in \mathbb{R}^{d}$. Intuitively $\mathbf{h}_u^L$ characterizes $u$ using a combination of its own features and features aggregated from its neighborhood. 
\looseness=-1
\vspace{-0.1in}

\noindent \\
\textbf{\textit{Scoring Partitions. }}
\looseness=-1 Recall from eq.~\ref{eq:state}, each partition at time $t$ is represented using the nodes belonging to that partition. Building upon this, we compute the score of each partition $p \in \CP^t$ with respect to the node $v$ selected in \textit{Phase 1} using all the nodes in  $p$.  In contrast to predicting a fixed-size score vector corresponding to number of partitions~\cite{nazi2019gap,tsitsulin2023graph}, the proposed design choice makes the model inductive to the number of partitions. Specifically, the number of partitions are not directly tied to the output dimensions of the neural model. 

 Having obtained the transformed node embeddings through a \gnn{}  in Eq.~\ref{eq:gnn}, we now compute the (unnormalized) score for node $v$ selected in phase 1 for each partition $p \in \CP^t$ as follows:
\vspace{-0.03in}
\begin{equation}
\label{eq:part_score}
\begin{aligned}    \operatorname{PartScore}&(p, v){=}  \operatorname{AGG} ( \{\operatorname{MLP}( \sigma( h_v|h_u) ) \\& \quad \quad \quad \forall u \in \mathcal{N}(v) \ni \operatorname{PART}(\CP^t,u) = p \}  )
\end{aligned}
\end{equation}

The above equation concatenates the selected node $v$'s embedding with its neighbors $ u \in \mathcal{N}(v)$ that belong to the partition $p$ under consideration. In general, the strength of a partition assignment to a node is higher if its neighbors also belong to the same partition. The above formulation surfaces this strength in the embedding space. The concatenated representation $(h_v|h_u) \forall u\in \mathcal{N}(v)$ is passed through an MLP that converts the vector into a score (scalar). We then apply an aggregation operator (e.g., mean) over all neighbors of $u$ belonging to $\CP^t$ to get an unnormalized score for partition $p$. Here $\sigma$ is an activation function.

To compute the normalized score at step $t$ is finally calculated as softmax over the all partitions $p \in \CP^t $ for the currently selected node $v$. Mathematically, the probability of taking action $a^t{=}p$ at time step $t$ at state $S^t$ is defined as:
\begin{equation}
\label{eq:select_part}
  \pi((a^t{=}p)/S^t) = \frac{\exp \left(\operatorname{PartScore}  \left( p,v \right) \right) }{\sum_{p^{\prime} \in \CP^t } \exp \left(\operatorname{PartScore}\left( p^{\prime}, v \right)\right)} 
\end{equation}

During the course of trajectory of length $T$, we sample action $a^t  \rev{\in \CP}$ i.e., the assignment of the partition for the node selected in phase 1 at step $t$ using policy $\pi$.

\noindent \\
\textbf{State Transition.}
After action $a^t$ is applied at state $S^t$, the state is updated to $S^{t+1}$ that involves updating the partition set $\CP^{t+1}$. Specifically, if node $v$ belonged to  $i^{th}$ partition at time $t$ and its partition has been changed to $j$ in phase 2, then we apply the below operations in order.
\begin{align}
\hspace{0.6in}
\label{eqn:state_update}
P^{t+1}_i \leftarrow P^{t}_i \setminus \{v\} \text{ and }
P^{t+1}_j \leftarrow P^{t}_j \cup \{v\} 
\vspace{-0.2in}
\end{align}
\subsection{Reward, Training \& Inference}
\textbf{Reward. }Our aim is to improve the value of the overall partitioning objective. One way is to define the reward $R^t$ at step $t\geq0$ as the change in objective value  of the partitioning i.e $ Obj(\CG, \CP^t)$ at step $t$. Specifically,\vspace{-.05in}
\begin{equation}
\label{eq:reward_scaled}
     R^t =  \frac{(Obj(\CG, \CP^{t}) -Obj(\CG, \CP^{t+1})) }{(Obj(\CG, \CP^{t})+Obj(\CG, \CP^{t+1}))}\cdot \lambda 
\end{equation}


 Here $\lambda$ is a hyperparameter that is used to scale the reward. The above reward expression incentivizes significant improvements in the objective function. This is achieved mathematically by considering both the change and the current value of the objective function. This design steers the model  towards prioritizing substantial improvements, especially in low objective function regions, ultimately guiding it towards the overall minimum.
 
 However, above definition of reward focuses on short-term improvements instead of long-term.  Hence, to prevent this local greedy behavior and to capture the combinatorial aspect of the selections, we use \textit{discounted rewards} $D^t$ to increase the probability of actions that lead to higher rewards in the long term~\cite{sutton2018reinforcement}. The discounted rewards are computed as the sum of the rewards over a \textit{horizon} of actions with varying degrees of importance (short-term and long-term). Mathematically,
\vspace{-0.05in}
   \begin{equation}
        \label{Eq_Disc_Returns}
    D^t=R^t +\gamma R^{t+1}+\gamma^{2}R^{t+2}+\ldots= \sum_{j=0}^{T-t} \gamma^{j} R^{t+j}
    \end{equation}\vspace{-0.00in}
    where $T$ is the length of the horizon and $\gamma \in (0,1]$ is a \textit{discounting factor} (hyper-parameter) describing how much we favor immediate rewards over the long-term future rewards. 

    The above reward mechanism provides flexibility to our framework to be versatile to objectives of different nature, that may or may not be differentiable. This is an advantage over existing neural methods~\cite{nazi2019gap, tsitsulin2023graph} where having a differentiable form of the partitioning objective is a pre-requisite. 

\noindent
\textbf{Policy Loss Computation and Parameter Update.} Our objective is to learn parameters of our policy network in such a way that actions that lead to an overall improvement of the partitioning objective are favored more over others.  Towards this, we use \textit{REINFORCE gradient estimator}~\cite{williams1992simple} to optimize the parameters of our policy network. Specifically, we wish to maximize the reward obtained for the horizon of length $T$ with discounted rewards $D^t$. Towards this end, we define a reward function $J(\pi_{\theta})$ as:
\vspace{-0.05in}
\begin{equation}
     J(\pi_{\theta}) =\mathbb{E} \big[ \sum_{t=0}^T \left( D^t \right)  \big]
 \end{equation}\vspace{-0.05in}

We, then, optimize $J(\pi_{\theta})$  \rev{via policy gradient~\cite{sutton2018reinforcement}} as follows:\vspace{-0.03in}
\begin{equation}
{
\nabla J(\pi_{\theta})= \left[ \sum_{t=0}^T \left( D^t  \right) \nabla_{\theta} log\pi_{\theta}({a}^t/S^t)  \right]}
\label{eq:loss_rl}
\end{equation}\vspace{-0.03in}

\noindent
\textbf{Training and Inference.} For a given graph, we optimize the parameters of the policy network $\pi_\theta$ for $T$ steps. Note that the trajectory length $T$ is not kept very large to avoid the long-horizon problem. During inference, we compute the initial node embeddings, obtain initial partitioning and then run the forward pass of our policy to improve the partitioning objective over time.


\vspace{-0.08in}
\subsection{Time Complexity}
\label{sec:time_compl}
The time complexity of {\namemodel} during inference is \\ $\mathcal{O}\big((\alpha \times \beta \rev{\times \vert \CE \vert}) +  ( \vert \CE \vert + k ) \times T' \big)$.  Here $\alpha$ is the number of anchor nodes, $\beta$ is number of random walk iterations, $k$ is the number of partitions and $T'$ is the number of iterations during inference. Typically $\alpha$, $\beta$ and $k$ are $<< \vert \CV \vert$ and $T'=o(\vert \CV \vert) $. Further, for sparse graphs $\vert \CE \vert =\mathcal{O}(\vert \CV \vert )$. Hence time complexity of {\namemodel} is $o(\vert \CV \vert^2)$. 
For detailed derivation please see Appendix ~\ref{derv:time_compl}.

%% file: 05_expts.tex
\section{Experiments}

In this section, we demonstrate the efficacy of {\namemodel} against state-of-the-art methods and establish that: 
\begin{itemize}

\item {\bf Efficacy and Robustness:} {\namemodel} produces the best results over diverse partitioning objective functions. This establishes the robustness of \namemodel.
\item {\bf Inductivity:} As one of the major strengths, unlike existing neural models such as {\gap}~\cite{nazi2019gap}, {\dmon}~\cite{tsitsulin2023graph} and {\mcpool}~\cite{bianchi2020spectral}, our method {\namemodel} is inductive on the number of partitions and consequently can generalize to unseen number of partitions. 
\end{itemize}
\textbf{Code: }Our code base is accessible at \url{https://github.com/idea-iitd/NeuroCut}.

\subsection{Experimental Setup}
\subsubsection{Datasets:} 
We use \textit{four real datasets} for our experiments. They are described below and their statistics are present in Table~\ref{tab:datasets}. For all datasets we use their largest connected component.
\begin{itemize}
    \item \looseness=-1 \textbf{Cora} and \textbf{Citeseer}~\cite{sen2008collective}: These are citation networks where nodes correspond to individual papers and edges represent citations between papers. The node features are extracted using a bag-of-words approach applied to paper abstracts.
    
    \looseness=-1

\item  \textbf{Harbin}~\cite{li2019learning}: This is a road network extracted from Harbin city, China. The nodes correspond to road intersections and node features represent latitude and longitude of a road intersection.

\item \textbf{Actor}~\cite{platonov2023critical}:
This dataset is based upon Wikipedia data where each node in the graph corresponds to an actor, and the edge between two nodes denotes co-occurrence on the same Wikipedia page. Node features correspond to keywords in Wikipedia pages. This is a heterophilous dataset~\cite{zheng2022graph} where nodes tend to connect to other nodes that are dissimilar.

\item \textbf{Facebook}~\cite{McAuley2012LearningTD}: In social network analysis, a $k$-way cut can delineate communities of users with minimal inter-community connections. This partitioning unveils hidden social circles and provides insights into how different groups interact within the larger network. The Facebook dataset, used for community detection, does not include node features. In this dataset, nodes represent users, and edges represent friendships.
\item \textbf{Stochastic Block Models (SBM)}~\cite{abbe2017community}: SBMs are synthetic networks and produce graphs with communities, where subsets of nodes are characterized by specific edge densities within the community. This allows us to validate our method by comparing the identified partitions with the ground truth communities. This dataset does not include node features.

\begin{table}[htbp]
\centering
\caption{Dataset Statistics}
\vspace{-0.13in}
\label{tab:datasets}
\scalebox{0.73}{
\begin{tabular}{ccclp{0.6in}p{0.7in}p{0.74in}}
\hline
\textbf{Dataset} & $\vert \mathcal{V} \vert$ & $\vert \mathcal{E} \vert$  & \textbf{\#Features} & \textbf{Average Degree} & \textbf{Clustering Coefficient} & \textbf{Degree Assortativity} \\ 
\midrule
Cora             & 2485              & 5069              & 1433                 & \textcolor{black}{4.07}                     & \textcolor{black}{0.23}                            & \textcolor{black}{-0.07}                         \\ 
CiteSeer         & 2120              & 3679              & 3703                 & \textcolor{black}{3.47}                     & \textcolor{black}{0.169}                           & \textcolor{black}{0.0075}                        \\ 
Harbin           & 6598              & 10492             & 2                    & \textcolor{black}{3.18}                     & \textcolor{black}{0.036}                           & \textcolor{black}{0.22}                          \\ 
Actor            & 6198              & 14879             & 931                  & \textcolor{black}{4.82}                     & \textcolor{black}{0.05}                            & \textcolor{black}{-0.048}                        \\
\rev{Facebook}         & \textcolor{black}{1034} & \textcolor{black}{26749} & \textcolor{black}{-} & \textcolor{black}{51.7} & \textcolor{black}{0.526} & \textcolor{black}{0.431} \\
\rev{SBM}              & \textcolor{black}{500} & \textcolor{black}{5150} & \textcolor{black}{-} & \textcolor{black}{20.6} & \textcolor{black}{0.18} & \textcolor{black}{-0.0059} \\ \hline
\end{tabular}
}
\end{table}\vspace{-0.1in}


\end{itemize}


\subsubsection{Partitioning objectives:} We evaluate our method on a diverse set of \textit{four partitioning objectives} described in Section ~\ref{sec:partitioning_obj}, namely \textit{Normalized Cut, Balanced Cut, k-MinCut, and Sparsest Cut}. In addition to evaluating on diverse objectives, we also choose a diverse number of partitions for evaluation, specifically,  $k=2,\; 5\; \text{and}\; 10$.

\subsubsection{Baselines:}
We compare our proposed method with both \textit{neural} as well as \textit{non-neural} methods. 

\textit{Neural baselines:} We compare with  {\dmon}~\cite{tsitsulin2023graph}, {\gap}~\cite{nazi2019gap},   {\mcpool} ~\cite{bianchi2020spectral} and \ortho~\cite{tsitsulin2023graph}. {\dmon} is the state-of-the-art neural attributed-graph clustering method. {\gap} is optimized for balanced normalized cuts with an end-to-end framework with a differentiable loss function which is a continuous relaxation version of normalized cut. {\mcpool} optimizes for normalized cut and uses an additional orthogonality regularizer. {\ortho} is the orthogonality regularizer described in {\dmon} and {\mcpool}. We also compare with DRL~\cite{gatti2022graph} which solves for Normalized cut at $k=2$ in Appendix~\ref{sec:comp_gatti} 

\textit{Non-neural baselines:} Following the settings of {\dmon}~\cite{tsitsulin2023graph}, we compare with {\kmeans} clustering applied on raw node features. \rev{We also compare with standard graph clustering methods hMetis~\cite{karypis1997multilevel} and Spectral clustering~\cite{ng2001spectral} in App Sec.~\ref{app:sec:non-neural-baselines}.}

\subsubsection{Other settings:} 


We run all our experiments on an Ubuntu 20.04 system running on Intel Xeon 6248 processor with 96 cores and 1 NVIDIA A100 GPU with 40GB memory for our experiments.
For {\namemodel} we used GraphSage\cite{hamilton2017inductive} as our GNN with number of layers $L=2$, learning rate as $ 0.0001$, hidden size = $ 32$. We used Adam optimizer for training the parameters of our policy network $\pi_\theta$.   For computing discounted reward in RL, we use discount factor $\gamma=0.99$. We set the length of trajectory $T$ during training as $2$. At time step $t$, the rewards are computed from time $t$ to $t+T$ and parameters of the policy $\pi_\theta$ are updated.  The default number of anchor nodes for computing positional embeddings is set to $35$. \rev{We set $\beta=100$ and $c=0.85$ for Eq.~\ref{eq:RWR_lipschitz}. We set scaling factor $\lambda=100$ in eq.~\ref{eq:reward_scaled}.}

\begin{table}[h!]
    \centering
      \caption{Results on Normalized Cut. Our model {\namemodel} produces the best (lower is better) performance across all datasets and the number of partitions $k$.}\vspace{-0.1in}
    \scalebox{0.82}{
    \begin{tabular}{ccccc}
    \toprule
    \small
    \textbf{Dataset} & \textbf{Method}  & \boldmath{$k=2$} & \boldmath{$k=5$} & \boldmath{$k=10$} \\ 
        \hline 
        
        \multirow{6}{*}{Cora}  & \kmeans  & 0.65  & 3.26  & 7.44  \\
                                & MinCutPool & 0.12 & 0.61 & 1.65 \\
                               & DMon  & 0.57  & 3.07  & 7.40  \\
                               & Ortho  & 0.80  & 1.88  & 4.06  \\
                               & GAP  & 0.10 & 0.68   & -  \\
                               
                                & \namemodel  & \textbf{0.02}  & \textbf{0.33}  & \textbf{0.92 } \\
        \hline
        \multirow{6}{*}{CiteSeer}
                                 & \kmeans  & 0.30	& 2.35 &	5.21  \\
                                & MinCutPool & 0.10 &	0.38 &	1.04 \\
                               & DMon  & 0.33	 &2.71	 & 6.84  \\
                               & Ortho  & 0.28	& 1.51 &	3.25  \\
                               & GAP  & 0.12  & -  & -  \\
                              
                                & \namemodel  & \textbf{0.02} &\textbf{0.20 }&	\textbf{0.44}  \\
        \hline
        \multirow{6}{*}{Harbin} & \kmeans  & 0.56	&2.34	 &5.40  \\
                                & MinCutPool & - & - & - \\
                               & DMon  & 0.98 &	3.25	 & - \\
                               & Ortho  &  - & - & -  \\
                               & GAP  & 0.25  & -  & -  \\
                                & \namemodel  & \textbf{0.01}&	\textbf{0.07	} &\textbf{0.28} \\
        \hline
        \multirow{6}{*}{Actor} & \kmeans  & 0.99  & 4.00  & 8.98  \\
                                & MinCutPool & 0.55 & 1.97 & 4.73 \\
                               & DMon  & 0.77  & 3.46  & 8.08  \\
                               & Ortho  & 1.05  & 3.98  & 8.90  \\
                               & GAP  & 0.20  & -  & -  \\
                                & \namemodel & \textbf{0.17} &	\textbf{0.99} &	\textbf{4.66} \\
        \bottomrule
    \end{tabular}     }
  
    \label{tab:norm_cut}
\end{table}

\begin{table}[h]
 \caption{Results on Sparsest Cut.  Our model {\namemodel} produces the best (lower is better) performance across all datasets and number of partitions $k$.}\vspace{-0.13in}
    \centering
    \scalebox{0.85}{
    \begin{tabular}{ccccc}
    \toprule
    \small
    \textbf{Dataset} & \textbf{Method}  & \boldmath{$k=2$} & \boldmath{$k=5$} & \boldmath{$k=10$} \\ 
        \hline\hline
        
        \multirow{6}{*}{Cora} 
                               & \kmeans  & 3.41 &	14.90 &	32.52 \\
                               & MinCutPool  & 0.52 & 2.44& 6.68 \\
                               & DMon  &0.45& 1.89& 5.82 \\
                               & Ortho  & 2.73& 7.06& 15.70 \\
                               & GAP  & 0.41 & 2.80 & -\\
                                & \namemodel  & \textbf{0.13}	 & \textbf{1.46} &	\textbf{3.03} \\
        \hline
        \multirow{6}{*}{CiteSeer}
                               & \kmeans  & 1.53	 & 13.70 &	22.20  \\
                               & MinCutPool  & 0.31& 1.32& 3.62 \\
                               & DMon  &0.35&1.21 & 3.62 \\
                               & Ortho & 0.88&4.15& 10.60  \\
                               & GAP  & 0.60 & - & -\\
                                & \namemodel  & \textbf{0.11} & \textbf{0.49}  &		\textbf{1.19}  \\
        \hline
        \multirow{6}{*}{Harbin} 
                               & \kmeans  & 1.91&	 7.44 &	17.03 \\
                               & MinCutPool  &-&-&- \\
                               & DMon  &2.01& -& - \\
                               & Ortho  &-&-&- \\
                               & GAP  & 1.56 & - & - \\
                                & \namemodel  & \textbf{0.06}  & \textbf{0.23}  & \textbf{0.82}  \\
        \hline
        \multirow{6}{*}{Actor} 
                               & \kmeans  & 5.43  & 19.40  & 40.34  \\
                               & MinCutPool  &2.44&8.51&19.81 \\
                               & DMon  &1.71&9.50&21.01 \\
                               & Ortho  &3.42& 16.55& 40.4 \\
                               & GAP  & 1.35 & - &  -\\
                                & \namemodel  & \textbf{0.65}  & \textbf{2.04}  & \textbf{2.88}  \\
        \bottomrule
    \end{tabular} }
   
    \label{tab:sparsest_cut}
\end{table}

\subsection{Results on Transductive Setting}

\looseness=-1
 \label{sec:res_transductive}
In the transductive setting, we compare our proposed method {\namemodel} against the mentioned baselines, where the neural models are trained and tested with the same number of partitions. Tables~\ref{tab:norm_cut}-\ref{tab:balanced_cut} present the results for different partitioning objectives across all datasets and methods. For the objectives under consideration, a smaller value depicts better performance. {\namemodel} demonstrates superior performance compared to both neural and non-neural baselines across four distinct partitioning objectives, highlighting its robustness. Further, we would also like to point out that in many cases, existing baselines produce \textit{``nan''} values which are represented by ``\textit{-}'' in the result tables. \rev{This is due to the reason that every partition is not assigned atleast one node which leads to the situation where denominator becomes $0$ in \textit{normalized},\textit{ balanced} and \textit{sparsest} cut objectives as defined in Sec.~\ref{sec:partitioning_obj}.}


\looseness=-1

 Our method {\namemodel} incorporates dependency during inference and this leads to robust performance. Specifically, as the architecture is auto-regressive, it takes into account the current state before moving ahead as opposed to a single shot pass in methods such as {\gap}~\cite{nazi2019gap}. Further, unlike other methods, {\namemodel} also incorporates positional information in the form of \textit{Lipschitz embedding} to better contextualize global node positional information.  We also observe that the non-neural method {\kmeans} fails to perform on all objective functions since its objective is not aligned with the main objectives under consideration.\vspace{-0.1in}

\begin{table}[t]
\vspace{-0.15in}
\caption{Results on k-MinCut. In most of the cases, our model {\namemodel} produces the best (lower is better) performance across all datasets and the number of partitions $k$. Values less than $10^{-2}$ are approximated to 0. }\vspace{-0.07in}
    \centering
    \scalebox{0.79}{
    \begin{tabular}{ccccc}
    \toprule
    \small
    \textbf{Dataset} & \textbf{Method}  & \boldmath{$k=2$} & \boldmath{$k=5$} & \boldmath{$k=10$} \\ 
        \hline \hline
        
        \multirow{6}{*}{Cora} 
                                & \kmeans  & 0.32	&0.34&	0.61  \\
                               & MinCutPool  & 0.06&0.12& 0.17 \\
                               & DMon  &0.05&0.09&0.14 \\
                               & Ortho  & 0.33&0.34& 0.38 \\
                               & GAP  & 0.05 &  0.10 & 0.11 \\
                                & \namemodel  & \textbf{ $\sim$ 0.0}	& \textbf{$\sim$ 0.0}& \textbf{0.06}  \\
        \hline
        \multirow{6}{*}{CiteSeer}
                                & \kmeans  & 0.12  & 0.27  & 0.43  \\
                               & MinCutPool  &0.04&0.08&0.10 \\
                               & DMon  &0.05&0.07&0.12 \\
                               & Ortho  &0.13&0.47&0.30 \\
                               & GAP  & 0.05 & 0.08 & 0.10\\
                                & \namemodel  & \textbf{0.01}	& \textbf{0.02} &	\textbf{0.03} \\
        \hline
        \multirow{6}{*}{Harbin} 
                                & \kmeans  & 0.28  & 0.46  & 0.54  \\
                               & MinCutPool  &\textbf{0.0}&\textbf{0.0}&\textbf{0.0} \\
                               & DMon  &0.10&0.01& $\sim$ 0.0 \\
                               & Ortho  &\textbf{0.0}&\textbf{0.0}&\textbf{0.0} \\
                               & GAP  & 0.01 & 0.01 & 0.01\\
                                & \namemodel  & $\sim$ 0.0  & 0.01  & 0.03  \\
        \hline
        \multirow{6}{*}{Actor} 
                                & \kmeans  & 0.48  & 0.54  & 0.80  \\
                               & MinCutPool  &0.25&0.35& 0.42\\
                               & DMon  &0.17&0.39&0.44 \\
                               & Ortho  &0.35&0.65& 0.83 \\
                               & GAP  & 0.09 & 0.12 &  0.26\\
                                & \namemodel  & \textbf{ $\sim$ 0.0 } & \textbf{$\sim$ 0.0}  & \textbf{$\sim$ 0.0}  \\
        \bottomrule
    \end{tabular} }
    \label{tab:kmin_cut}
    \vspace{-0.08in}
\end{table}

\begin{table}[h]

\caption{Results on Balanced Cut. In most cases, our model {\namemodel} produces the best (lower is better) performance across all datasets and number of partitions $k$.}\vspace{-0.1in}
    \centering
    \scalebox{0.84}{
    \begin{tabular}{ccccc}
    \hline
    \small
    \textbf{Dataset} & \textbf{Method}  & \boldmath{$k=2$} & \boldmath{$k=5$} & \boldmath{$k=10$} \\ 
        \hline
        \multirow{6}{*}{Cora} 
                                & \kmeans  & 0.68  & 3.90  & 7.44  \\
                               & MinCutPool  & 0.13& \textbf{0.60}& 1.62 \\
                               & DMon  &0.11&0.48&1.47 \\
                               & Ortho  &0.80&1.89& 4.06 \\
                               & GAP  & \textbf{0.10} & 0.74 & - \\
                                & \namemodel  & 0.45	& 0.64
 &	\textbf{1.08} \\
        \hline
        \multirow{6}{*}{CiteSeer}
                                & \kmeans  & 0.42  & 2.64  & 5.30  \\
                               & MinCutPool  &0.09&0.38&1.04 \\
                               & DMon  &0.10&0.37&1.1 \\
                               & Ortho  &0.28&1.51&3.32 \\
                               & GAP  & 0.23 & - &  -\\
                                & \namemodel  &\textbf{0.07}	& \textbf{0.24}
 &	\textbf{0.60} \\
        \hline
        \multirow{6}{*}{Harbin} 
                                & \kmeans  & 0.56  & 2.37  & 5.41  \\
                               & MinCutPool  &-&-&- \\
                               & DMon  &1.30&-&- \\
                               & Ortho  &-&-&- \\
                               & GAP  & 0.72  & - &  -\\
                                & \namemodel  & \textbf{0.21}  & \textbf{0.11}  & \textbf{0.27}  \\
        \hline
        \multirow{6}{*}{Actor} 
                                & \kmeans  & 1.00  & 4.06  & 9.08  \\
                               & MinCutPool  &0.55&1.96&4.71 \\
                               & DMon  &0.34&2.01&4.80 \\
                               & Ortho  &1.05&3.90&8.91 \\
                               & GAP  & \textbf{0.25} & - & - \\
                                & \namemodel  & 0.59  & \textbf{1.69}  & \textbf{4.42}  \\
        \hline
    \end{tabular} }
    
    \label{tab:balanced_cut}
\end{table}

\subsection{Results on Inductivity to Partition Count}

\begin{table}[h!]
 \caption{Inductivity to unseen partition count. Here we set the target number of partitions $k=10$. I stands for inductive and T for transductive setting. In this table, the transductive version of {\namemodel} trained on $k=10$ serves as a point of reference when assessing the quality of the inductive version of {\namemodel}. }\vspace{-0.08in}
    \centering
    \scalebox{0.84}{
    \begin{tabular}{ccccccc}
    \hline
    \textbf{Dataset} & \textbf{Method}  & \boldmath{Normalized} & \boldmath{Sparsest} & \boldmath{k-MinCut} & \boldmath{Balanced} \\ 
        \hline
        
        \multirow{3}{*}{Cora} 
                                    & \kmeans  &  7.44  & 32.52  & 0.61 & 7.53  \\
                            & \namemodel-T & 0.92  & 3.03  & 0.06 & 1.08  \\
                                & \namemodel-I & 1.18  & 5.04  & 0.02 & 2.03  \\

                            \cline{1-6}
          \multirow{3}{*}{CiteSeer} 
                               & \kmeans  & 5.21  & 22.20  & 0.43 & 5.30  \\
                            
                            & \namemodel-T & 0.44  & 1.19  & 0.03 & 0.60  \\
                                & \namemodel-I & 0.63  & 1.62 & 0.03 & 0.62 \\
                            \cline{1-6}
      \multirow{3}{*}{Harbin} 
                                & \kmeans  & 5.40  & 17.03  & 0.54 & 5.41  \\
                            
                            & \namemodel-T & 0.28  & 0.82  & 0.03 & 0.27  \\
                                & \namemodel-I & 0.27  & 0.87  & 0.03 & 0.29  \\

        \hline
    \end{tabular}}
   
    \label{tab:inductive_partition_count}\vspace{-0.1in}
\end{table}

\begin{figure*}[t!]
\vspace{-0.2in}
     \centering
     \subfloat[][Normalized Cut]{\includegraphics[scale=0.25]{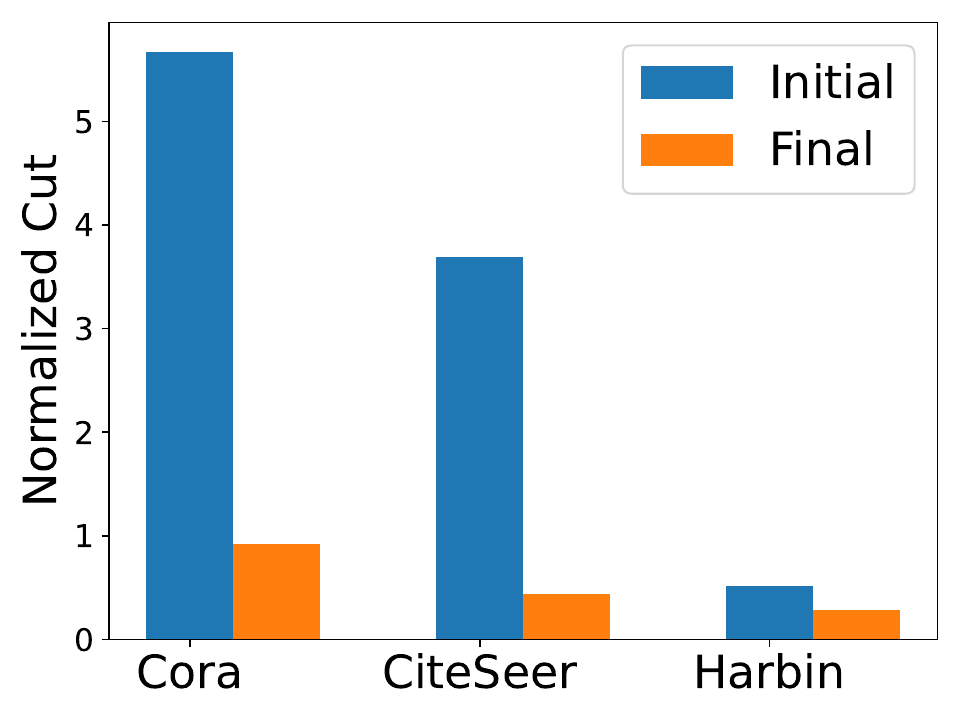}\label{fig:subfig1}}\hspace{0.1in}
     \subfloat[][Sparsest Cut]{\includegraphics[scale=0.25]{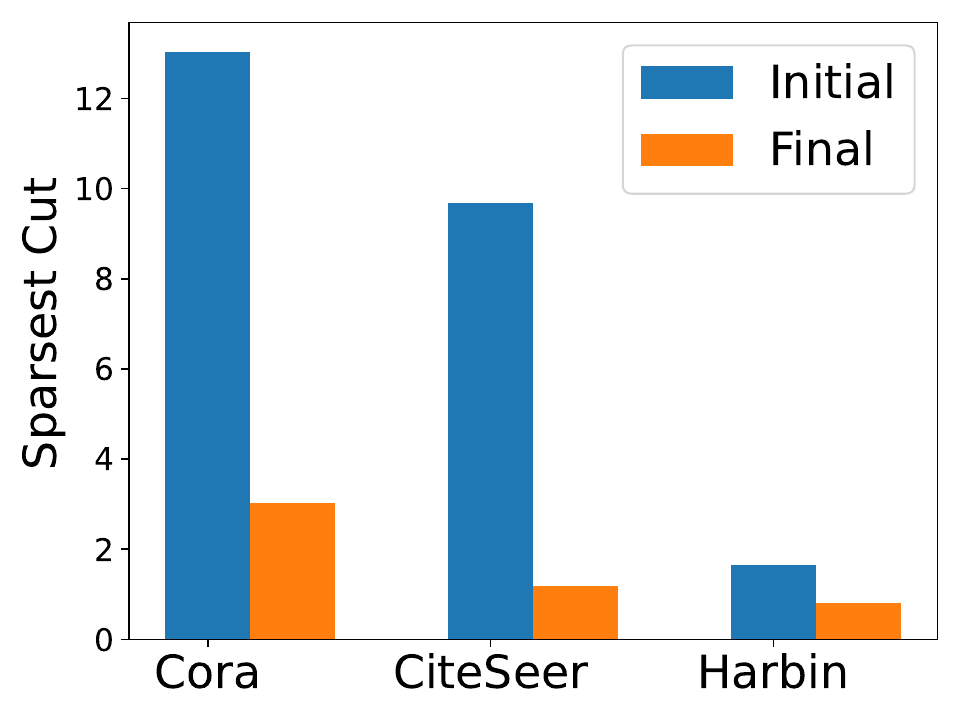}\label{fig:subfig1}}  \hspace{0.1in}
      \subfloat[][Balanced Cut]{\includegraphics[scale=0.25]{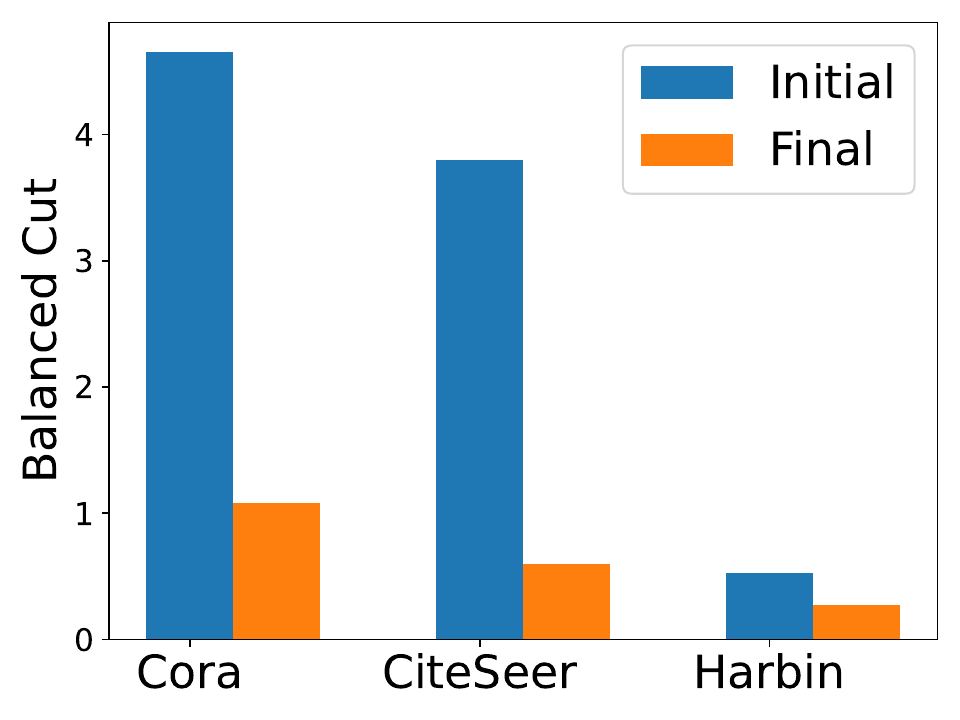}\label{fig:subfig3}}\hspace{0.05in}
     \subfloat[][k-MinCut]{\includegraphics[scale=0.25]{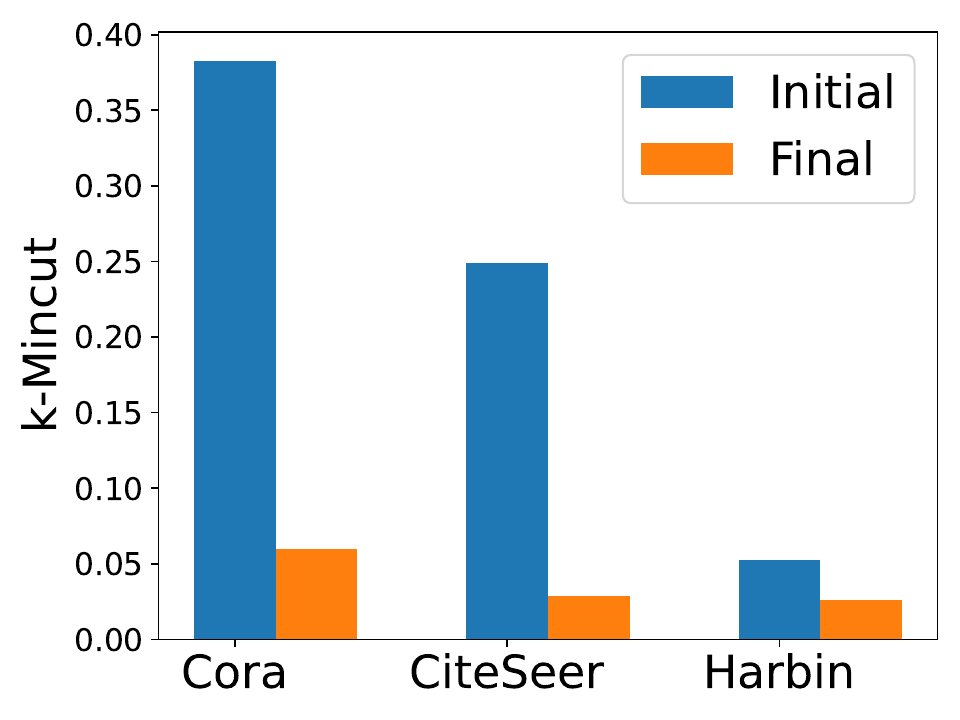}\label{fig:subfig4}}\vspace{-0.15in}
     \caption{ Results on the initial warm start and the final cut values obtained by {\namemodel} at $k{=}10$. It shows that our neural model {\namemodel} (Final) performs more accurate node and partition selection to optimize the objective function. Subsequently, there is a significant difference between the initial and final cut values.}
     \label{fig:initial_vs_final}
     \vspace{-3mm}
\end{figure*}

\looseness=-1
As detailed in Section~\ref{sec:phase2_node_selection}, the decoupling of parameter size and the number of partitions allows {\namemodel} to generalize effectively to an unseen number of partitions. In this section, we empirically analyze the generalization performance of {\namemodel} to an unknown number of partitions. We compare it with the non-neural baseline {\kmeans}, as the neural methods like {\gap}, {\dmon}, {\ortho}, and {\mcpool} cannot be employed to infer on an unseen number of partitions. In Table~\ref{tab:inductive_partition_count}, we present the results on inductivity. Specifically, we trained {\namemodel} on different partition sizes ($k=5$ and $k=8$) and tested it on an unseen partition size $k=10$. The inductive version, referred to as {{\namemodel}-I}\footnote{For clarity purposes, in this experiment we use {\namemodel}-T to refer to the transductive setting whose results are presented in Sec. ~\ref{sec:res_transductive}}, obtains high-quality results on different datasets.  Note that while the non-neural method such as {\kmeans} needs to be re-run for the unseen partitions, {\namemodel} only needs to perform  forward pass to produce the results on an unseen partition size.

\subsection{Ablation Studies}
\textbf{Initial Warm Start vs Final Cut Values.} {\namemodel} takes a warm start by partitioning nodes based on clustering over their raw node features and Lipschitz embeddings, which are subsequently fine-tuned auto-regressively through the \textit{phase 1} and \textit{phase 2} of {\namemodel}. In this section, we measure, how much the partitioning objective has improved since the initial clustering? Figure \ref{fig:initial_vs_final} sheds light on this question. Specifically, it shows the difference between the initial cut value after clustering and the final cut value from the partitions produced by our method. We note that there is a significant difference between the initial and final cut values, which essentially shows the effectiveness of {\namemodel}.

\noindent

\begin{figure}[t!]
     \centering
     \includegraphics[scale=0.34]{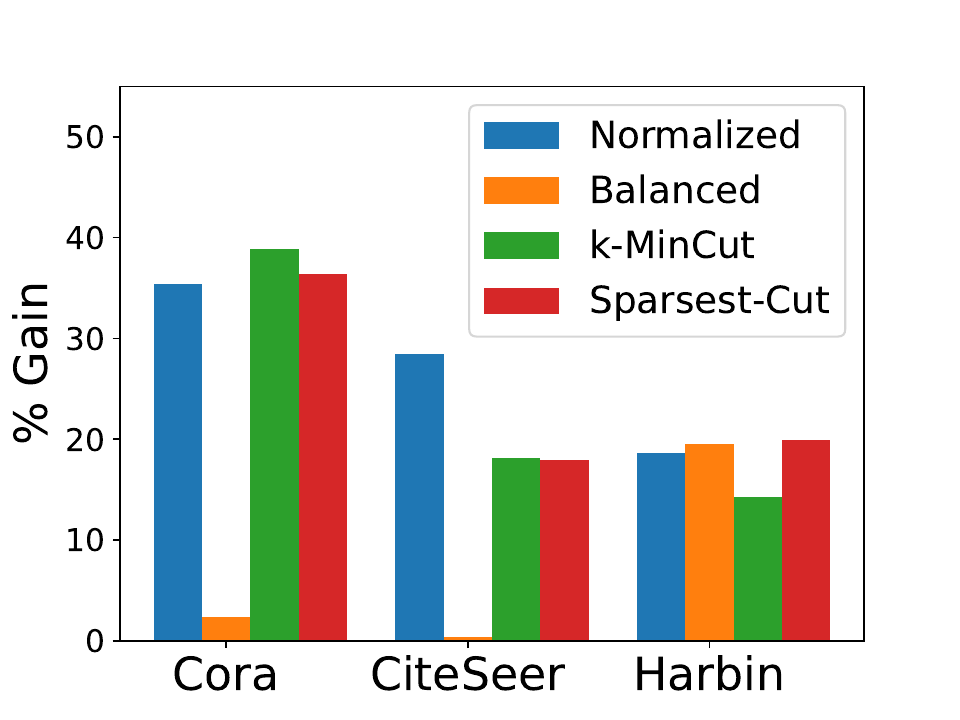}
     \vspace{-0.13in}
    \caption{Node Selection in \textit{Phase 1} with  Heuristic vs Random: Relative \% improvement (gain) in cut values obtained by {\namemodel} when using \textit{ different node selection strategies} for $k=5$. In most cases, our heuristic finds significantly better cuts than a random node selection procedure.}
     \label{fig:node_selection}
     \vspace{-0.2in}
\end{figure}

\textbf{Impact of Node Selection Procedures.} In this section, we explore the effectiveness of our proposed node selection heuristic in enhancing overall quality. Specifically, we measure the relative performance gain(in $\%$) obtained by {\namemodel} when using the node selection heuristic as proposed in Section ~\ref{sec:phase1_node_selection} over a random node selection strategy. \textcolor{black}{In our experiments for the random node selection strategy, we selected the best performing run across multiple seeds.} Figure \ref{fig:node_selection} shows the percentage gain for three datasets.  We observe that a simpler node selection where we select all the nodes one by one in a random order, yields substantially inferior results in comparison to the heuristic proposed by us. This suggests that the proposed sophisticated node selection strategy plays a crucial role in optimizing the overall performance of {\namemodel}. In Appendix ~\ref{app:node_selection_detailed}, we conduct a more detailed analysis of the observations presented in Figure \ref{fig:node_selection}, with particular attention to the variation in gains related to the Balanced Cut objective.

\textcolor{black}{\textbf{Impact of Clustering Initialization in
Warm-start Phase:} \textcolor{black}{To understand the importance of different initializations in the warm-up phase(\S~\ref{sec:initialization}), in Appendix ~\ref{sec:app:clus_impact} we perform an ablation study using three different initialization schemes, namely {\kmeans}(default),
density-based clustering  DBSCAN~\cite{ester1996density}, and Random initialization.}
}

\subsection{Impact of Cluster Strength on Performance}
In Section~\ref{app:cluster_strength} in appendix, we analyze the impact of cluster strength in networks on the performance of different methods.

%% file: 06_Conclusion.tex
\section{Conclusion}
In this work, we study the problem of graph partitioning with node features. Existing neural methods for addressing this problem require the target objective to be differentiable and necessitate prior knowledge of the number of partitions. In this paper, we introduced {\namemodel}, a framework to effectively address the graph partitioning problem with node features. {\namemodel} tackles these challenges using a reinforcement learning-based approach that can adapt to any target objective function. Further, attributed to its decoupled parameter space and partition count,  {\namemodel} can generalize to an unseen number of partitions. The efficacy of our approach is empirically validated through an extensive evaluation on four datasets, four graph partitioning objectives and diverse partition counts. Notably, our method shows significant performance gains when compared to the state-of-the-art techniques, proving its competence in both inductive and transductive settings.

\noindent\textbf{Limitations and Future Directions:} Achieving a sub-quadratic computational complexity with an inductive neural method for attributed graph partitioning is an open challenge. In {\namemodel}, node selection and perturbation are performed in a sequential fashion. One direction to improve the efficiency of {\namemodel} could be batch processing the selection and perturbation of multiple independent nodes simultaneously.   
Another interesting direction could be designing explanations of the neural methods---such as \namemodel---for graph combinatorial problems \cite{kakkad2023survey}.

%% file: 08_Ack.tex
\section{Acknowledgement}

We thank Kartik Sharma for the helpful suggestions. Rishi Shah acknowledges that the funding for the conference was provided by the CMU GSA Conference Funding. Sahil Manchanda acknowledges financial support by GP Goyal grant of IIT Delhi and Qualcomm Innovation Fellowship India. 

%% file: 07_appendix.tex
\appendix

\section{Appendix}

\subsection{Clustering for Initialization}
\label{app:clustering}
As discussed in sec~\ref{sec:initialization} in main paper, we first cluster the nodes of the graph into $k$ clusters where $k$ is the number of partitions.  Towards this, we apply {\kmeans} algorithm~\cite{macqueen1967some} on the nodes of the graph where a node is represented by its raw node features and positional representation i.e $ \mathbf{emb_{init}}(u) \; \forall u \in \CV $ based upon eq.~\ref{eq:emb_init}. Further, we used $L_{\inf}$ norm as the distance metric for clustering.

\subsection{Time Complexity Analysis of \namemodel}
\label{derv:time_compl}

\textbf{(1)} First the positional embeddings for all nodes in the graph are computed. This involves running RWR for $\alpha$ anchor nodes for $\beta$ iterations (Eq. ~\ref{eq:RWR_lipschitz}). This takes $\mathcal{O}(\alpha \times \beta \rev{\times \vert \CE \vert})$ time.   \textbf{(2)} Next, GNN is called to compute embeddings of node. In each layer of GNN a node $v \in \CV$ aggregates message from $d$ neighbors where $d$ is the average degree of a node. This takes  $\mathcal{O}(\CV)$ time.  This operation is repeat for $L$ layers. Since $L$ is typically 1 or 2 hence we ignore this factor. \textbf{(3)} The node selection algorithm is used that computes score for each node based upon its neighborhood using eq.~\ref{eqn:node_selection}. This consumes $\mathcal{O}(\vert \CV \vert \times d)$ time. \textbf{(4)} Finally for the selected node, its partition has to be determined using eq.~\ref{eq:part_score} and ~\ref{eq:select_part}. This takes $\mathcal{O}( d + k)$ time, as we consider only the neighbors of the selected node to compute partition score.   Steps 2-4 are repeated for $T'$ iterations.  Hence overall running time is $\mathcal{O}\big((\alpha \times \beta \rev{\times \vert \CE \vert} ) +  ( \vert \CV \vert \times d  + k ) \times T' \big)$. Typically $\alpha$, $\beta$ and $k$ are $<<\vert  \CV \vert $ and $T'=o(\vert \CV \vert) $. Since $\vert \CV \vert \times d \approx \vert \CE \vert $, hence complexity is $o( \vert \CE \vert   \times \vert \CV \vert \big)$   Further, for sparse graphs $\vert \CE \vert =\mathcal{O}(\vert \CV \vert )$. Hence time complexity of {\namemodel} is $o(\vert \CV \vert^2 )$.

\subsection{Time Complexity Comparison}
\rev{Table~\ref{tab:time_complexity}  presents the complexities of {\namemodel} and other prominent neural and non-neural baselines algorithms.  In the table, $\vert \mathcal{V}  \vert$, $\vert \mathcal{E}  \vert$, $k$, $d$ are the number of nodes, edges, partitions, and average node degree respectively.
While some neural algorithms exhibit faster complexity, they require separate training for each partition size ($k$). In contrast, {\namemodel}, once trained, can generalize to any value. Consequently, for practical workloads, {\namemodel} presents a more scalable option in terms of computation overhead and storage (one model versus separate models for each).}

\begin{table}[t]
\caption{Time complexity of different methods.\label{tab:time_complexity}}
\centering
\begin{tabular}{lll}
\hline
\textcolor{black}{\textbf{Method}}       & \textcolor{black}{\textbf{Complexity}}            & \rev{\textbf{Comments}}                          \\ \hline
\textcolor{black}{DMon}         & $\textcolor{black}{\mathcal{O}(d^2{\times} \vert \mathcal{V}  \vert + \vert \mathcal{E}\vert )}$ & \rev{Per iteration complexity} \\ 
\textcolor{black}{MinCutPool}   & $\textcolor{black}{\mathcal{O}(d^2{\times} \vert \mathcal{V}  \vert + \vert \mathcal{E}\vert )}$ & \rev{Per iteration complexity} \\ 
\textcolor{black}{Ortho}        & $\textcolor{black}{\mathcal{O}(\vert \mathcal{V} \vert{\times} k^2  )}$ &   \rev{Per iteration complexity}                     \\ 
\textcolor{black}{GAP}          & $\textcolor{black}{\mathcal{O}(\vert \mathcal{V} \vert{\times}k^2 + \vert \mathcal{V} \vert^2 )}$ & \rev{Per iteration complexity} \\ 
\textcolor{black}{Spectral}     & $\textcolor{black}{\mathcal{O}(\vert \mathcal{V}  \vert ^3 )}$                    &      \rev{Inference complexity}         \\ 
\textcolor{black}{{\namemodel}}     & $\textcolor{black}{o(\vert \mathcal{V} \vert^2 )}$                             &    \rev{Inference complexity}    \\ \hline
\end{tabular}
\vspace{-3mm}
\end{table}

\rev{The quadratic time complexity of {\namemodel} may pose challenges for very large graphs. However, this complexity remains faster than spectral clustering, a widely used graph partitioning algorithm. Also, for the baseline neural methods(DMon, MinCutPool, Ortho and GAP), the time complexity provided is per iteration. The number of iterations often ranges between 1000 to 2000 and there is no clear understanding of how it varies as a function of the graph (like density, diameter, etc.)}

\subsection{Comparison: Non-neural Baselines}
\label{app:sec:non-neural-baselines}

\rev{
We compare the performance of our method against graph based clustering algorithms hMETIS and Spectral Clustering and partition algorithm Gomory-Hu Tree~\cite{gomory1961multi}. Since these methods are not compatible with datasets having raw node features, we compare with two datasets namely Facebook~\cite{leskovec2012learning} and Stochastic Block Model(SBM)~\cite{abbe2017community} which don't have node features for this comparision. Table~\ref{tab:comparison_non_neural_featureless} compares the performance of {\namemodel} against hMETIS and Spectral Clustering. The details of these datasets are presented in Table~\ref{tab:datasets}. In addition to non-neural methods, we also show the performance of neural methods MinCutPool, DMon, Ortho and GAP on these datasets.} 

\rev{The performance of {\namemodel} is better than non-neural methods hMETIS, Spectral and Gomory-Hu Tree in most of the cases on Facebook dataset. Further, on SBM dataset, it matches the performance of hMETIS and Spectral clustering. \textcolor{black}{The SBM dataset has a clear community structure, making it an easy instance for partitioning algorithms, resulting in similar performance across different methods.} We would also like to highlight that in the case of $k$$-$$mincut$, Gomory-Hu Tree algorithm generated trivial partitions for these datasets where almost all nodes were assigned the same partition. The other neural methods such as MinCutPool, DMon, Ortho, and GAP fail to produce a valid solution in most of the cases. This is possibly because these baselines are not robust to perform on datasets without raw node features. Overall, {\namemodel} outperforms the baselines on diverse objectives and datasets.}

\begin{table}[t]
\centering
\caption{\textcolor{black}{Performance of different methods on Facebook and SBM dataset. Lower values are better. $'-'$ denotes \textit{nan}.}}
\vspace{-2mm}
\label{tab:comparison_non_neural_featureless}
\scalebox{0.77}{
\begin{tabular}{llcccc}
\midrule
\small
\multirow{2}{*}{\textbf{\textcolor{black}{Dataset}}} & \multirow{2}{*}{\textbf{\textcolor{black}{Method}}} & \multicolumn{4}{c}{\textbf{\textcolor{black}{Metrics}}} \\
                                  &                                      & \textbf{\textcolor{black}{Normalized}} & \textbf{\textcolor{black}{Sparsest}} & \textbf{\textcolor{black}{Balanced}} & \textbf{\textcolor{black}{$k$-MinCut}} \\ \hline
\multirow{7}{*}{\textbf{\textcolor{black}{Facebook}}} & \textbf{\textcolor{black}{{\namemodel}}}                             & \textbf{\textcolor{black}{0.257}}                   & \textbf{\textcolor{black}{4.121}}                  & \textbf{\textcolor{black}{0.972}}                  & \textcolor{black}{0.015}            \\  
                                  & \textbf{\textcolor{black}{hMETIS}}                               & \textcolor{black}{1.15}                    & \textcolor{black}{52.319}                 & \textcolor{black}{1.115}                  & \textcolor{black}{0.200}            \\  
                                  & \textbf{\textcolor{black}{Spectral}}                             & \textcolor{black}{1.67}                    & \textcolor{black}{4.72}                   & \textcolor{black}{2.459}                  & \textcolor{black}{0.003}            \\ 
                                  & \textbf{\textcolor{black}{Gomory-Hu Tree}}                             & \textcolor{black}{4.00}                    & \textcolor{black}{5.00}                   & \textcolor{black}{4.792}                  & \textbf{\textcolor{black}{$\sim$0}}            \\ 
                                  
                                  & \textbf{\textcolor{black}{MinCutPool}}                           & \textcolor{black}{-}                       & \textcolor{black}{-}                      & \textcolor{black}{-}                      & \textcolor{black}{0.023}            \\  
                                  & \textbf{\textcolor{black}{DMon}}                                 & \textcolor{black}{-}                       & \textcolor{black}{-}                      & \textcolor{black}{-}                      & \textbf{\textcolor{black}{$\sim$0}}          \\  
                                  & \textbf{\textcolor{black}{Ortho}}                                & \textcolor{black}{-}                       & \textcolor{black}{-}                      & \textcolor{black}{-}                      & \textcolor{black}{0.05}             \\  
                                  & \textbf{\textcolor{black}{GAP}}                                  & \textcolor{black}{-}                       & \textcolor{black}{-}                      & \textcolor{black}{-}                      & \textcolor{black}{0.025}            \\ \hline
\multirow{7}{*}{\textbf{\textcolor{black}{SBM}}}      & \textbf{\textcolor{black}{NeuroCUT}}                             & \textcolor{black}{0.191}                   & \textcolor{black}{3.939}                  & \textcolor{black}{0.191}                  & \textcolor{black}{0.038}            \\  
                                  & \textbf{\textcolor{black}{hMETIS}}                               & \textcolor{black}{0.191}                   & \textcolor{black}{3.939}                  & \textcolor{black}{0.191}                  & \textcolor{black}{0.038}            \\  
                                  & \textbf{\textcolor{black}{Spectral}}                             & \textcolor{black}{0.191}                   & \textcolor{black}{3.939}                  & \textcolor{black}{0.191}                  & \textcolor{black}{0.038}            \\  
                                  & \textbf{\textcolor{black}{Gomory-Hu Tree}}                             & \textcolor{black}{4.003}                    & \textcolor{black}{48.75}                   & \textcolor{black}{4.787}                  & \textcolor{black}{{0.0075}}            \\ 
                                  & \textbf{\textcolor{black}{MinCutPool}}                           & \textcolor{black}{-}                       & \textcolor{black}{-}                      & \textcolor{black}{-}                      & \textcolor{black}{0.0}              \\  
                                  & \textbf{\textcolor{black}{DMon}}                                 & \textcolor{black}{-}                       & \textcolor{black}{-}                      & \textcolor{black}{-}                      & \textcolor{black}{0.0}              \\  
                                  & \textbf{\textcolor{black}{Ortho}}                                & \textcolor{black}{-}                       & \textcolor{black}{-}                      & \textcolor{black}{-}                      & \textcolor{black}{$\sim$0}          \\  
                                  & \textbf{\textcolor{black}{GAP}}                                  & \textcolor{black}{-}                       & \textcolor{black}{-}                      & \textcolor{black}{-}                      & \textcolor{black}{0.035}            \\ \hline
\end{tabular}
}
\vspace{-3mm}
\end{table}

\subsection{Comparison with  Gatti et al.~\cite{gatti2022graph}}
\label{sec:comp_gatti}
In Table~\ref{tab:gatti}, we compare the performance of our method against DRL method proposed by Gatti et al.~\cite{gatti2022graph} which solves the normalized cut problem for the case where the number of partitions is exactly two.  We observe that {\namemodel} outperforms DRL on all datasets.

\begin{table}[t]
\vspace{-1mm}
\centering
\caption{Gatti et al. on Normalized Cut  at $k=2$.\label{tab:gatti} }
\scalebox{0.85}{
\begin{tabular}{lcccccc}
\toprule
$\dfrac{\text{\bf Method}\rightarrow}{\text{\bf Dataset}\downarrow}$& \textbf{Cora} & \textbf{CiteSeer} & \textbf{Harbin} & \textbf{Actor} & \textbf{SBM} & \textbf{Facebook} \\
\midrule
NeuroCUT    & \textbf{0.02} & \textbf{0.02} & \textbf{0.01} & \textbf{0.17} & \textbf{0.191} & \textbf{0.257} \\
Gatti et al. & 0.355 & 0.29 & 0.13 & 1.00 & 0.72 & 0.92 \\
\bottomrule
\end{tabular}
}
\vspace{-3.5mm}

\end{table}

\subsection{Impact of Clustering Initialization}

\label{sec:app:clus_impact}

\rev{
To understand the importance of different initializations in the warm-up phase, we perform an ablation study on $2$ datasets namely Cora and CiteSeer using 3 different initialization namely {\kmeans}, density-based clustering \textit{DBSCAN}~\cite{ester1996density}, and Random initialization. In Table~\ref{tab:warm_up_clustering} we observe the performance on normalized cut at $k=5$. We observe that {\kmeans} performs the best in this experiment. The improvement observed when using {\kmeans} or \textit{DBSCAN} over Random shows that a good initialization i.e. warm-up does help in improving quality of partitions. In DBSCAN we set $eps=0.9$ and $min\_samples=2$. It is worth noting that DBSCAN's performance can vary based on the parameter selection. Nonetheless, the primary goal of this experiment is to demonstrate the advantageous role of a good initialization, specifically in contrast to random initialization.
}
\begin{table}[t]
    \caption{Impact of Clustering initialization in Warm-up Phase of {\namemodel} in the normalized cut objective at $k=5$.}
    \centering
    \small
    \color{black} 
    \begin{tabular}{lccc}
        \toprule
        \textbf{Dataset} & \textbf{\kmeans} & \textbf{Random} & \textbf{DBSCAN} \\
        \midrule
        Cora & \textbf{0.33} & 0.79 & 0.55 \\
        CiteSeer & \textbf{0.20} & 0.51 & 0.70 \\
        \bottomrule
    \end{tabular}
    \label{tab:warm_up_clustering}
    \vspace{-3mm}
\end{table}
\begin{table}[t]
\centering
\caption{Statistics for Normalized Cut (3 runs) for $k=5$}
\label{tab:norm_cut_appendix}
\scalebox{0.75}{
\begin{tabular}{lcccc}
\hline
\textbf{Dataset} & \textbf{Our Heuristic} & \textbf{Random Selection} & \textbf{Random Selection} & \textbf{\% Gain} \\
                 &                        & \textbf{Avg}              & \textbf{Best}             & \textbf{(from best)} \\
\midrule
Cora     & 0.33 $\pm$ 0.0024  & 0.461 $\pm$ 0.0164  & 0.445 & 35 \\
Citeseer & 0.20 $\pm$ 0.00961 & 0.276 $\pm$ 0.0177  & 0.258 & 29 \\
Harbin   & 0.073 $\pm$ 0.0027 & 0.097 $\pm$ 0.0131  & 0.086 & 18 \\
\hline
\end{tabular}
}
\end{table}

\begin{table}[t]
\vspace{-1mm}
\centering
\caption{Statistics for Balanced Cut (across 3 runs) for $k=5$}
\label{tab:bal_cut_appendix}
\scalebox{0.74}{
\begin{tabular}{lcccc}
\hline
\textbf{Dataset} & \textbf{Our Heuristic} & \textbf{Random Selection} & \textbf{Random Selection} & \textbf{\% Gain} \\
                 &                        & \textbf{Avg}              & \textbf{Best}             & \textbf{(from best)} \\
\midrule

Cora     & 0.642 $\pm$ 0.0065 & 0.753 $\pm$ 0.0574 & 0.672  & 2   \\
Citeseer & 0.248 $\pm$ 0.0049 & 0.307 $\pm$ 0.039  & 0.252  & 0.5 \\
Harbin   & 0.1087 $\pm$ 0.0010 & 0.1313 $\pm$ 0.011  & 0.1293 & 19  \\
\hline
\end{tabular}
}
\vspace{-2mm}
\end{table}

\begin{table}[h]
\centering
\caption{Number of Nodes Re-assigned after Initial Assignment by NeuroCUT on the Balanced Cut objective.}
\label{tab:nodes_reassigned}
\vspace{-2mm}
\scalebox{.85}{
\begin{tabular}{lcccc}
\hline
\textbf{Dataset} & \textbf{Nodes Perturbed} & \textbf{Total Nodes} & \textbf{Percentage} \\
\midrule
Cora     & 791  & 2495  & 31.7 \\
Citeseer & 413  & 2120  & 19.4 \\
Harbin   & 187  & 6598  & 2.8  \\
\hline
\end{tabular}
}
\vspace{-6mm}
\end{table}

\textcolor{black}{\subsection{Impact of Node Selection Procedures}
\label{app:node_selection_detailed}
\rev{In Fig.~\ref{fig:node_selection} in main paper, random selection performs well only in Balanced Cuts. The objective in Balanced Cuts (Eq.~\ref{eq:balanced_cuts}.) is not only a function of the combined weight of cut edges, but also balancing the number of nodes across partitions. This additional node-balancing term is not present in the objective functions of the other cut definitions. Due to this reduced importance of optimizing the cut value, random does well, since even when an incorrect node is selected, it can still be utilized to keep the partition sizes balanced.  Moreover, we find that when the warm-up assignment is less accurate and the percentage of nodes re-assigned by NeuroCUT to a different partition is high, a random selection of nodes is closer in efficacy to our heuristic selection. In contrast, in Harbin, the percentage of nodes perturbed to a new partition is significantly smaller. In this scenario, the probability that random selection will select this small subset is lower and hence amplifying the effectiveness gap between random and heuristic selection.}
}

\subsection{Impact of Cluster Strength on Performance}
\label{app:cluster_strength}
\rev{To understand the impact of community structure on performance of {\namemodel}, we generate Stochastic Model Block(SBM) graphs with different intra cluster strength. In Table~\ref{table:sbm},  we observe the performance of different methods on the normalized cut objective at $k=5$.} \rev{The baseline methods hMETIS and Spectral Clustering perform worse when the strength within communities is low indicated by Intra Cluster Edge Probability and Clustering Coefficient values. Although {\namemodel} outperforms or matches existing methods in all cases, however, the gap between baselines and {\namemodel} increases significantly when the community structure is not strong. Further, the neural baselines DMon, MinCutPool and Ortho generated `nan' values for this experiment.}

\begin{table}[h]
\centering
\vspace{-2mm}
\small
\caption{Performance on SBM Dataset}
\label{table:sbm}
\scalebox{0.7}{
\begin{tabular}{>{\centering\arraybackslash}p{0.24in}>{\centering\arraybackslash}p{0.65in}>{\centering\arraybackslash}p{0.73in}>{\centering\arraybackslash}p{0.2in}>{\centering\arraybackslash}p{0.2in}|>{\centering\arraybackslash}p{0.5in}>{\centering\arraybackslash}p{0.3in}>{\centering\arraybackslash}p{0.5in}}
\hline
\textbf{SBM} & \multicolumn{4}{c}{\textbf{Dataset Statistics}} & \multicolumn{3}{c}{\textbf{Normalized Cut}} \\ \hline
& \textbf{Intra Cluster Edge Probability} & \textbf{Clustering Coefficient} & $\vert \CV \vert$ & $\vert \CE \vert$ & {\namemodel} & hMETIS & Spectral Clustering \\ \hline
1 & 0.005 & 0.003 & 495 & 555 & \textbf{0.3089} & 0.331 & 0.379 \\ 
2 & 0.01 & 0.0029 & 525 & 661 & \textbf{0.3434} & 0.5889 & 0.4259 \\ 
3 & 0.2 & 0.184 & 500 & 5150 & 0.1912 & 0.1912 & 0.1912 \\ 
4 & 0.4 & 0.3822 & 500 & 10102 & 0.1153 & 0.1153 & 0.1153 \\ \hline
\end{tabular}
}
\vspace{-2mm}
\end{table}

\subsection{Impact of $\beta$ on \namemodel}
\rev{We study the impact of $\beta$ parameter(eq. 6) which is the number of iterations in random walk with restart. In Table~\ref{tab:beta_impact} we present  the performance of {\namemodel} using different $\beta$ on normalized-cut objective at $k=5$. We observe that {\namemodel} improves with more iterations as $\beta$ increases and then its performance stabilizes.} 

\begin{table}[h]
\vspace{-2mm}
\caption{Impact of $\beta$ parameter on {\namemodel} \label{tab:beta_impact} on normalized cut at $k=5$ on the Cora dataset.}
\centering
\color{black}
\begin{tabular}{cc}
\hline
$\beta$ & Normalized Cut \\
\hline
1 & 2.12 \\
3 &  1.57 \\
10 & 0.33 \\
50 & 0.33 \\
100 & 0.33 \\
\hline
\end{tabular}
\vspace{-2mm}
\end{table}

\subsection{Stability Across Multiple Runs}

Table~\ref{tab:std_dev_neurocut} shows the Mean and Standard deviation of {\namemodel} for Normalized cut at $k=5$ for 3 runs. The Std Dev. is lower than the Mean showing the stability of {\namemodel} across multiple runs.

\begin{table}[h]
    \centering
    \caption{Mean and Standard Deviation for different datasets on Normalized Cut at $k=5$ for \namemodel}
    \begin{tabular}{|l|c|}
        \hline
        \textbf{Dataset} & \textbf{Mean $\pm$ Std Dev} \\
        \hline
        Cora & $0.337 \pm 0.0024$ \\
        Citeseer & $0.20 \pm 0.00961$ \\
        Harbin & $0.073 \pm 0.0027$ \\
        Actor & $0.97 \pm 0.0389$ \\
        \hline
    \end{tabular}
    
    \label{tab:std_dev_neurocut}
\end{table}